\documentclass[10pt,letterpaper]{article}
\usepackage[OT1]{fontenc}
\usepackage{lmodern}
\usepackage[letterpaper,top=24mm,bottom=25mm,left=25mm,right=25mm]{geometry}
\usepackage{microtype}
\usepackage[dvipsnames]{xcolor}
\usepackage{graphicx}
\usepackage{amsmath,amssymb,bm}
\usepackage{booktabs,array,multirow}
\usepackage{enumitem}
\usepackage{float}
\usepackage[section]{placeins}
\usepackage{titlesec}
\usepackage{fancyhdr}
\usepackage{caption,subcaption}
\usepackage[numbers,sort&compress]{natbib}
\usepackage[hyphens]{url}
\usepackage{hyperref}
\usepackage[capitalize,noabbrev,nameinlink]{cleveref}

\definecolor{accent}{HTML}{254B60}
\definecolor{abstractbg}{HTML}{F6F8F9}
\definecolor{abstractborder}{HTML}{D7E1E5}
\hypersetup{colorlinks=true,allcolors=accent,
  pdftitle={Playfulness as Feedback for Agentic Game Development},
  pdfauthor={},pdfsubject={Agentic game development},pdfkeywords={}}
\setlist[itemize]{leftmargin=1.4em,itemsep=3pt,topsep=4pt}
\setlist[enumerate]{leftmargin=1.6em,itemsep=3pt,topsep=4pt}
\titleformat{\section}{\large\sffamily\bfseries\color{accent}}{\thesection}{0.7em}{}
\titleformat{\subsection}{\normalsize\sffamily\bfseries\color{accent}}{\thesubsection}{0.7em}{}
\titleformat{\subsubsection}{\normalsize\bfseries}{\thesubsubsection}{0.7em}{}
\titleformat{\paragraph}[runin]{\normalsize\bfseries}{}{0pt}{}
\titlespacing*{\section}{0pt}{14pt}{6pt}
\titlespacing*{\subsection}{0pt}{10pt}{4pt}
\titlespacing*{\paragraph}{0pt}{6pt}{0.6em}
\makeatletter
\renewcommand{\maketitle}{%
  \begin{center}
    \noindent\makebox[\linewidth][s]{%
      \includegraphics[height=11mm]{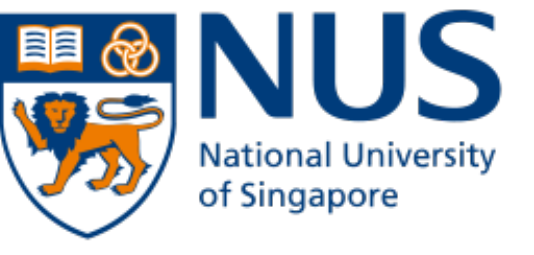}%
      \hfill
      \includegraphics[height=11mm]{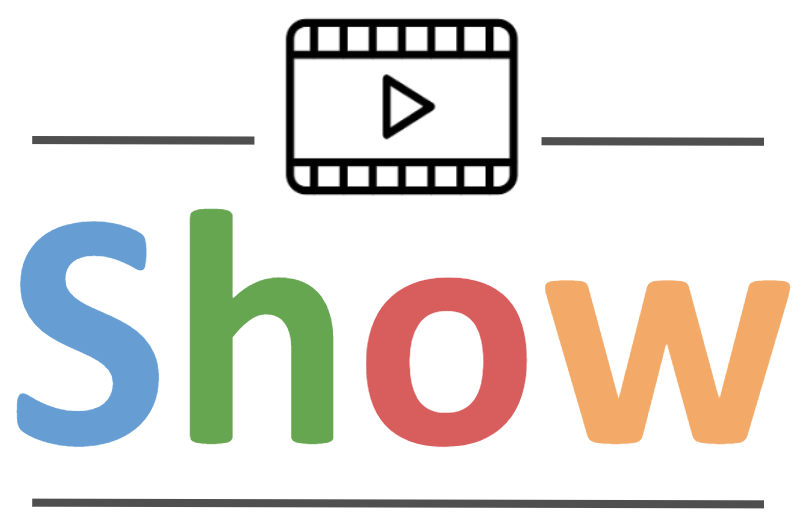}%
    }\par
    \vspace{8pt}
    \rule{\linewidth}{0.5pt}\par
    \vspace{8pt}
    {\fontsize{21}{25}\selectfont\sffamily\bfseries\color{accent}\@title\par}
    \vspace{8pt}
    \rule{\linewidth}{0.5pt}\par
    \vspace{10pt}
    {\normalsize\@author\par}
  \end{center}
  \vspace{8pt}
}
\makeatother
\newsavebox{\abstractbox}
\renewenvironment{abstract}{%
  \begin{lrbox}{\abstractbox}\begin{minipage}{\dimexpr\linewidth-24pt-2\fboxrule\relax}
  \small\setlength{\parskip}{4pt}
  {\sffamily\bfseries\color{accent}Abstract\par}\medskip
}{%
  \end{minipage}\end{lrbox}
  \begingroup\setlength{\fboxsep}{12pt}
  \noindent\fcolorbox{abstractborder}{abstractbg}{\usebox{\abstractbox}}
  \endgroup\par\vspace{8pt}
}

\title{Playfulness as Feedback for Agentic Game Development}
\author{%
  Weijia Mao\textsuperscript{1}, Kevin Qinghong Lin\textsuperscript{1},
  Zhen Fang\textsuperscript{2}, Danze Chen\textsuperscript{1},
  Dongxing Mao\textsuperscript{1}, Mike Zheng Shou\textsuperscript{1,*}\\[-1pt]
  {\small \textsuperscript{1}Show Lab, National University of Singapore\\[-1pt]
   \textsuperscript{2}University of Science and Technology of China\\[-1pt]
   * Corresponding author}
}
\date{}
\begin{document}
\maketitle
\begin{abstract}
Coding agents can generate runnable games from natural-language descriptions. However, a runnable game does not guarantee an excellent gameplay experience for players. In our view, a playful game should provide 1) reliable core functionality, 2) appealing visual design and assets, and 3) rich gameplay that supports exploration and progression. How agents can improve these aspects of playfulness remains underexplored. In our work, we propose \textbf{Play2Evolve}, a recursive game development framework with three stages: Initialization, Refinement, and Evolution. Using playfulness as feedback, agents playtest the game and diagnose issues in its functionality, visual presentation, and gameplay. Their feedback guides bug fixes and asset improvements. It also helps agents propose new gameplay and aesthetic goals for continued game evolution. To enrich game materials, we combine a video generation model with code to create appealing static and dynamic assets and integrate them into interactive gameplay. Finally, Play2Evolve achieves state-of-the-art performance among the evaluated methods, scoring 60.05 on GameCraft-Bench compared with 50.81 after Game Initialization and 52.46 for Play2Code. Human overall-quality ratings reach 3.67, compared with 2.38 and 2.43, respectively. Our code will be released at \url{https://github.com/showlab/Play2Evolve}.
\par\medskip
{\footnotesize
\setlength{\parskip}{2pt}
\noindent\textbf{Correspondence:} Mike Zheng Shou at
\href{mailto:mike.zheng.shou@gmail.com}{\texttt{mike.zheng.shou@gmail.com}}\par
\noindent\textbf{Project Page:}
\url{https://showlab.github.io/Play2Evolve/}\par
}

\end{abstract}

\section{Introduction}
\label{sec:intro}
\begin{figure}[htbp]
\centering
\includegraphics[width=\textwidth,trim=0 48bp 0 48bp,clip]{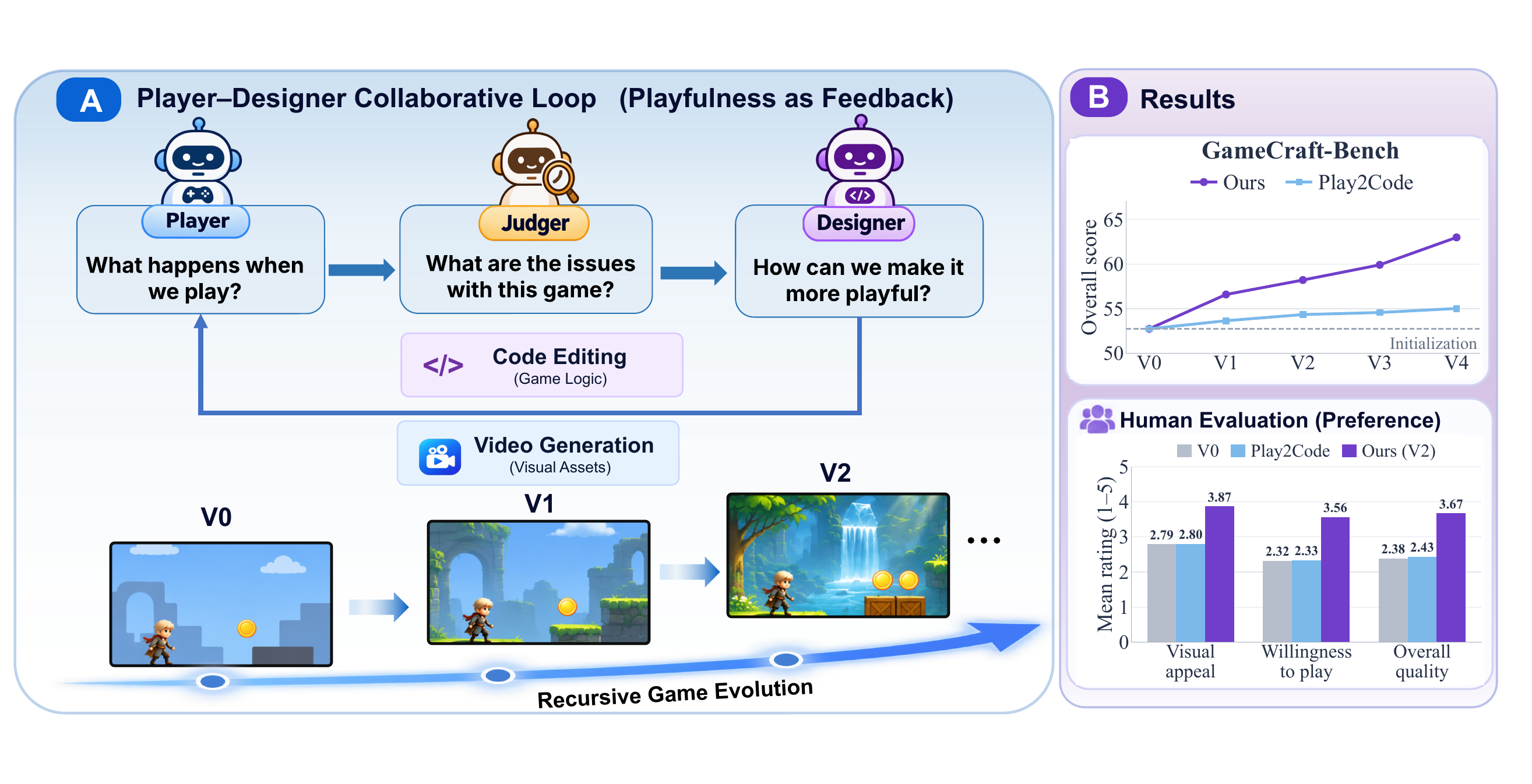}
\captionsetup{hypcap=false,skip=4pt}
\caption{Overview of Play2Evolve. (A) Player, Judge, and Designer agents use playtest feedback to guide code and video generation, turning runnable prototypes into playful games through defect repair, richer visual assets, and new gameplay. (B) GameCraft-Bench~\cite{luo2026gamecraft} scores and human ratings improve over initialization and Play2Code~\cite{huang2026play2code}.}
\captionsetup{hypcap=true}
\label{fig:teaser}
\end{figure}

Recently, coding agents~\cite{openai2025codex,anthropic2025claudecode,glm2025glm45,deepseek2025v32} have advanced rapidly and have been applied to a broad range of software development tasks, including code generation, editing, and debugging. 
This progress also extends to game development. GameGPT~\cite{chen2025gamegpt}, AutoUE~\cite{yin2026autoue}, and OpenGame~\cite{jiang2026opengame} show that agents can build runnable game prototypes from high-level specifications, such as natural-language game descriptions. Play2Code~\cite{huang2026play2code} further uses gameplay feedback to revise generated games. Together, these works demonstrate substantial progress in agentic game development.

However, creating games that players want to play remains a challenge for agentic game development. Existing work can produce runnable prototypes, but these games may still suffer from limited gameplay variety and overly simple visual assets that weaken the player experience. These limitations lead us to ask: what aspects of a game matter most to players seeking a better gaming experience? In our view, a \emph{playful} game should meet at least three requirements for a positive player experience: i) complete and reliable core functionality that enables responsive interaction and uninterrupted gameplay, ii) appealing visual design and assets that attract players and support immersion in the game world, and iii) rich gameplay that supports exploration and progression. Much existing work on agentic game development prioritizes the first requirement. Support for enriching visual assets and developing more diverse gameplay remains limited.

Therefore, we propose \textbf{Play2Evolve, a recursive game development framework}. Our goal is to enable agents to create and continually develop games that players find engaging and playful. Specifically, we draw on recursive self-improvement framework to let agents playtest games, identify problems, and propose revisions. Their reviews focus on functional correctness, visual quality, and gameplay richness. We apply recursive improvement to two aspects of game development: 1) repairing core functionality, where playtest feedback guides bug fixes and visual asset checks against existing requirements. 2) upgrading gameplay and aesthetics, where feedback guides new functions, more diverse gameplay, and visual improvements. Agents playtest each revised game again to verify changes and inform the next iteration. To create visually appealing games, we combine video generation with code for adaptive visual asset management. The video model produces static and dynamic assets, while code integrates them with player input, interactions, and game state.

Our method consists of three stages: Game Initialization, Game Refinement, and Game Evolution. \textbf{i) Game Initialization} combines code and video generation to create a runnable game from the user specification. \textbf{ii) Game Refinement} assesses observed gameplay against the original requirements and uses playtest feedback to repair functional and visual defects. It coordinates code and asset changes to address the issues revealed during interaction. \textbf{iii) Game Evolution} uses agent playtest recordings of the previous version as the basis for developing the next version. An agent reviews these recordings together with the game requirements to propose upgrades to visual assets and functionality. The proposed upgrades are implemented and evaluated through further playtests, checking both new goals and existing gameplay. This process can repeat over multiple iterations, with each retained version supplying the playtest evidence for its successor.

In our experiments, we evaluate game quality on two benchmarks, GameCraft-Bench~\cite{luo2026gamecraft} and GameASG-Bench~\cite{zhang2026gameasg}. Besides automated evaluation, we conduct a human study to assess perceived playability, visual quality, and willingness to play. Our upgraded games outperform the initialization versions on both benchmarks and in the human study, and exceed Play2Code across all four GameCraft-Bench dimensions and all five human evaluation dimensions. Further experiments show continued gains across successive evolution cycles. We also compare agent playtesting strategies and examine how Behavior Skill improves exploration, highlighting the trade-off between coverage and resource use.

\noindent Our contributions are as follows:
\begin{enumerate}
  \renewcommand{\labelenumi}{\arabic{enumi})}
  \item We frame playful game development around reliable core functionality, visual appeal, and rich gameplay that supports exploration and progression. Play2Evolve organizes these goals into a recursive framework that builds, repairs, and upgrades games through play.
  \item We develop a play-guided method that explores gameplay, repairs evidenced defects, and proposes new gameplay and aesthetic goals. An A/B Judge compares game versions using playtest evidence to verify improvements while checking retained functionality.
  \item We integrate video generation with code for adaptive game asset management to create visually appealing games. Gameplay feedback guides asset generation and revision, while code integrates the materials into interactive gameplay.
  \item Play2Evolve achieves state-of-the-art performance on GameCraft-Bench, with a mean Overall score of 60.05 compared with 52.46 for adapted Play2Code. Human overall-quality ratings reach 3.67, compared with 2.43 for Play2Code and 2.38 after Game Initialization.
\end{enumerate}

\section{Related Work}
\label{sec:related}

\noindent\textbf{Agentic game development and evaluation.} General coding agents and multi-agent systems support planning, implementation, and testing~\cite{anthropic2025claudecode,openai2025codex,google2025geminicli,glm2025glm45,hong2024metagpt,qian2024chatdev}. Long-horizon studies examine how earlier edits affect later development~\cite{zeng2026multistepswe}. Game-focused systems combine agents, runtimes, and reusable project knowledge~\cite{chen2025gamegpt,yin2026autoue,jiang2026opengame,hu2026vibegame}. Benchmarks assess complete projects through gameplay, runtime checks, and visual evidence~\cite{chi2026gamedevbench,luo2026gamecraft,zhang2026gameasg,chen2026gamexpert,chen2026swegame,dong2026cegamebench,che2026gamelogicbench}. Iterative feedback and verification also appear in general agent workflows~\cite{madaan2023selfrefine,shinn2023reflexion,jolicoeur2025avr,yan2026hoh}. Play2Code connects GUI playtesting with code revision~\cite{huang2026play2code}. Concurrent RSIGame tracks local repairs and global quality~\cite{wu2026rsigame}, while RGC uses executable play policies and experience-oriented review~\cite{chen2026recursive}. Player experience and automated playtesting motivate evaluating more than build success~\cite{sweetser2005gameflow,yannakakis2011experience,ferdous2021search,ferdous2025curiosity,tan2024cradle}. Executable policies and feedback-driven skill development offer ways to explore interactive environments~\cite{liang2023codeaspolicies,wang2024voyager,ma2024eureka}. Play2Evolve uses play evidence to guide both requirement repair and repeatable game upgrades.

\noindent\textbf{Video generation for interactive content.} Video models provide motion from text or visual conditions~\cite{guo2024animatediff,blattmann2023stablevideo,yang2025cogvideox}. Interactive world models, including Genie, GameNGen, GameGen-X, GameFactory, and Matrix-Game, generate game video conditioned on player actions~\cite{bruce2024genie,valevski2024gamengen,che2025gamegenx,yu2025gamefactory,zhang2025matrixgame}. Matrix-Game 4.0 also couples executable game rules with a video renderer to produce playable worlds~\cite{zhang2026matrixgame4}. Structured state and executable code can guide video synthesis~\cite{zhan2026magpie,wu2026wizard,li2026videococo,chen2025code2video,liang2025videoagent}. Other work preserves gameplay entities during generative rendering or turns videos into reusable game materials~\cite{liu2026anygamerenderer,zhou2026gamecreationagents}. Play2Evolve generates reusable materials at different scopes and integrates them into an editable game project. Play feedback then guides revisions to both materials and gameplay code.

\section{Method}
\label{sec:method}
\begin{figure}[htbp]
  \centering
  \captionsetup{skip=2pt}
  \includegraphics[width=\textwidth,trim=0 30bp 0 0,clip]{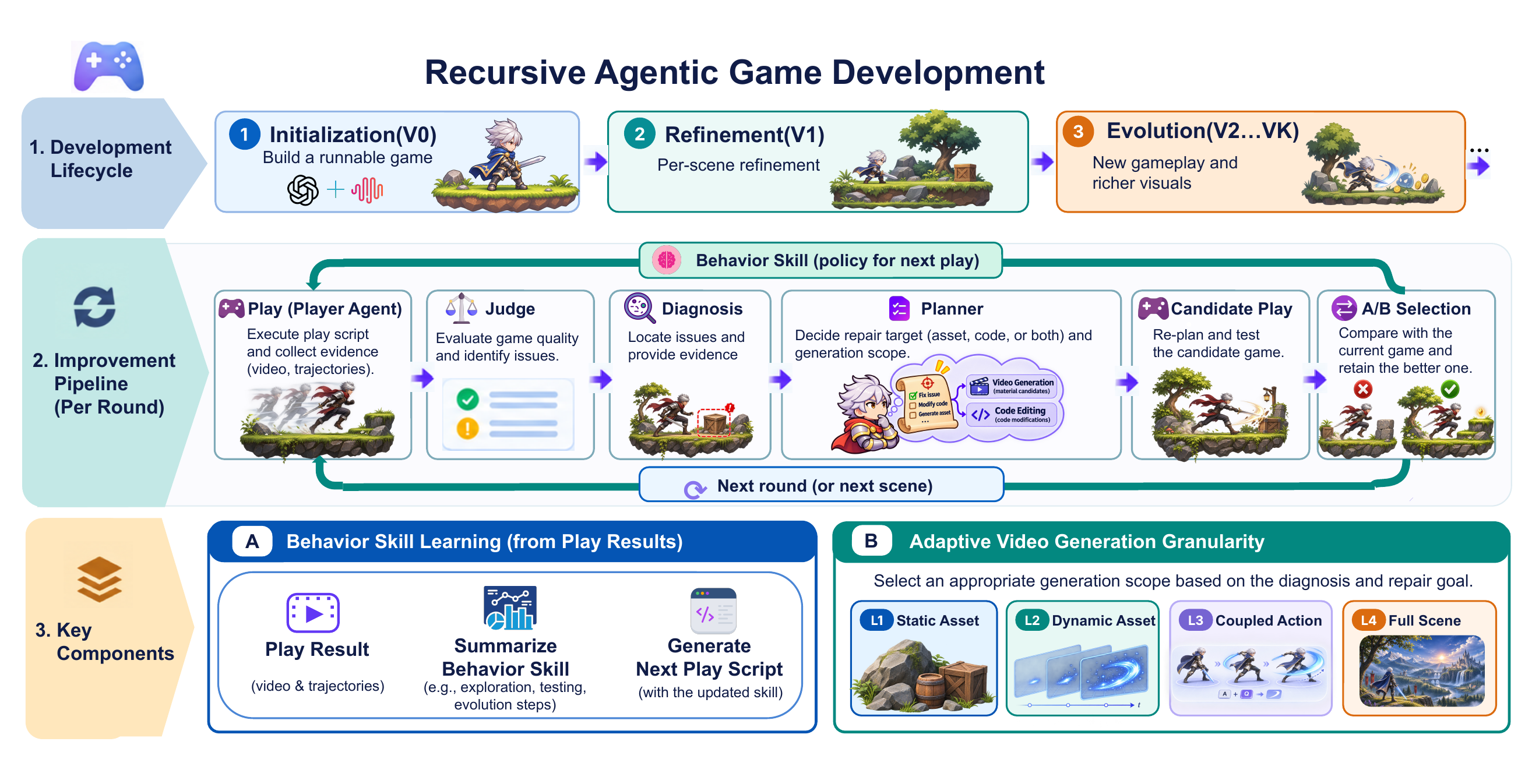}\par
  \caption{Overview of Play2Evolve's recursive game development pipeline. (1) Initialization builds a runnable game with code and video generation. Playtest feedback guides Refinement to repair functional and visual defects, and Evolution to develop new gameplay and aesthetic improvements. (2) Each round turns agent play into diagnosis, planning, video and code revisions, candidate play, and A/B selection. (3) Behavior Skills guide later play, while adaptive generation selects static, single-subject, coupled-action, or full-scene materials.}
  \label{fig:pipeline}
\end{figure}

Play2Evolve begins with Game Initialization (\hyperref[sec:method-bootstrap]{Sec.~\ref*{sec:method-bootstrap}}), which creates a runnable game with video-generated materials. Game Refinement (\hyperref[sec:method-initialization]{Sec.~\ref*{sec:method-initialization}}) uses agent-authored play programs to find and repair defects. Game Evolution (\hyperref[sec:method-rsi]{Sec.~\ref*{sec:method-rsi}}) then plans game-specific improvements and tests them through agent play. The retained game can enter further evolution cycles. Adaptive video generation can revise materials during initialization and both loops. Our recursive pipeline is shown in Fig.~\ref{fig:pipeline}.

Let $x$ denote the public game specification, $P$ an executable game project, $\mathcal{A}$ its material plan, and $\mathcal{T}$ its declared replay traces. We name agent roles by their responsibilities. Several roles may use the same underlying model.

\subsection{Game Initialization}
\label{sec:method-bootstrap}
An independent Requirement Extractor reads the public specification $x$. A Builder Agent creates the runnable game, input adapter, material plan, and replayable acceptance traces. It can interleave code edits with video generation requests and inspect generated frames before integration. We summarize the output as
\begin{equation}
\label{eq:bootstrap}
(P^{\mathrm{init}},\mathcal{A},\mathcal{T})
=\mathcal{B}(x;\pi_{\mathrm{build}},V_{\theta}),
\end{equation}
where $\pi_{\mathrm{build}}$ is the Builder Agent and $V_{\theta}$ is the video generation model.

\paragraph{Video and code collaboration.}
The Builder Agent assigns each visual role to one of 4 generation levels. \textbf{1) Static material} provides a still sprite or background. \textbf{2) Single dynamic material} animates one actor or prop. \textbf{3) Coupled-object material} generates interacting subjects and props in one synchronized clip. \textbf{4) Full-scene material} supplies motion across the environment. The agent chooses a level based on the required motion and interaction, together with the control needed during play. Each request specifies the visual subject, motion, prompt, and, for dynamic material, a visual reference.

The material plan also assigns responsibility between video and code. Video supplies appearance and the selected motion, while code places the material in the game, triggers or switches clips, and retains player input, collision, progress, outcomes, and interface state. For coupled actions, the interacting subjects belong to one clip rather than being drawn twice. For a full-scene clip, the interface and interactive state remain under code control. This division lets later repairs change a material's level without giving up gameplay control.

Initialization checks game launch, material bindings, and replay traces. The resulting runnable project $P^{\mathrm{init}}$ forms the \textrm{V0} checkpoint and enters Game Refinement.

\subsection{Game Refinement}
\label{sec:method-initialization}
Game Initialization provides a runnable package, but gameplay functions, visual assets, and interactions may still be incomplete or incorrect. The refinement loop uses agent play to expose these defects before repairing them. The Player seeks broad coverage of gameplay branches and deeper progress toward scenario goals, so diagnosis can draw on more than the entry scene.

Each refinement round follows a recursive improvement cycle of play, diagnosis, repair, and verification. A Player Agent plans goal-oriented playtest groups from the public requirements and game code. It writes a reactive play program for each group, then runs the program to collect actions, frames, and trajectories. Since refinement aims to satisfy the existing requirements, the Quality Judge checks this evidence for missing or incorrect core functionality and basic defects in visual materials. A Diagnosis Agent compares the findings with the requirements and source-material previews to locate their causes. It attributes each evidenced issue to generated materials, code, or both.

\paragraph{Video generation level selection.}
Based on the diagnosis, a Unified Planner groups issues by their repair boundary and proposes a repair at the current generation level. A malformed actor or an incorrect pose can be regenerated without changing the generation scope or motion type. If the source material is correct but its placement, scale, or playback timing is wrong, the plan changes code instead. These repairs retain the current division of responsibility between video and code.

The planner also proposes a higher-level alternative when observed defects require additional motion or joint generation. For example, a static actor sliding through an action can gain articulated motion through a new animation. Contact, synchronization, or occlusion errors between separately rendered subjects and props can motivate a coupled clip, such as generating a character and rope together. Each alternative specifies how code preserves input, collision, and state transitions. If the required interaction cannot be represented by finite clips while preserving these controls, the planner retains the current scope.

The Builder Agent combines these proposals into at most 2 complete-game candidates, one at the current levels and one with the justified level changes. It generates the requested materials and edits the code for each candidate. The Player Agent adapts the play program to each version. An A/B Judge compares their recorded play with the incumbent and retains a candidate only when the repair is verified without critical regression.

For playtest group $j$ and repair round $r$, this loop is
\begin{equation}
\label{eq:initialization}
\begin{aligned}
e_{j,r} &= \operatorname{Play}(P_{j,r},\pi_{j,r}),\\
\mathcal C_{j,r} &= \operatorname{Revise}_{\mathrm{levels}}(P_{j,r},x,e_{j,r}),\\
\mathcal E_{j,r} &= \{\operatorname{Play}(Q,\pi_Q):Q\in\mathcal C_{j,r}\},\\
P_{j,r+1} &= \operatorname{Select}_{\mathrm{refine}}(P_{j,r},\mathcal C_{j,r},\mathcal E_{j,r}).
\end{aligned}
\end{equation}
Here $\pi_{j,r}$ is the selected play program, $\mathcal C_{j,r}$ contains at most 2 complete-game candidates, and $\pi_Q$ is adapted to candidate $Q$. The evidence set $\mathcal E_{j,r}$ records their plays. The selected game and program advance together. Selection retains the incumbent when a repair is unverified or regresses a previously satisfied requirement.

\paragraph{Behavior Skill updates.}
In program-based refinement, a Skill Agent summarizes recorded play into a game-local textual Behavior Skill. A changed, nonempty skill guides the Player when it writes the next play program and can be passed to the evolution loop. The evolution loop continues this play-to-skill update after its own rounds. The update changes the Player's context, not model weights. For example, a saved Skill for Gravity Shift instructs the Player to change gravity when the ball is blocked at an edge, and to undo an unhelpful move before trying another direction.

Each planned playtest group allows up to 3 repair rounds. The retained game and play program pass to the next group. We denote the final refined project by $P^{\mathrm{refine}}$ (\textrm{V1}) and evaluate frozen checkpoints separately with GameCraft-Bench.

\subsection{Game Evolution}
\label{sec:method-rsi}
Game Evolution recursively develops a game through playtest feedback. It extends Game Refinement (Sec.~\ref{sec:method-initialization}) in two aspects. 1) It recursively develops successive game versions. Each cycle uses playtest videos of the retained previous version and the game requirements as the basis for the next upgrade. The Designer first proposes new gameplay and aesthetic goals, and the Builder implements an upgraded version before the iterative playtest and improvement loop begins. The retained version and its playtest evidence then support the next cycle, allowing development to continue beyond a single upgrade. 2) The judging objective shifts from repairing defects against existing requirements to improving gameplay functionality and aesthetics. Judges assess whether the upgraded game delivers these improvements while preserving working functionality.

We describe the evolution protocol used in our successive-evolution runs. Let $P_k$ be the retained game before evolution cycle $k$, with $P_0=P^{\mathrm{refine}}$ denoting \textrm{V1}. At the start of cycle $k$, a Player Agent performs a fresh playtest of the retained previous version $P_k$. We denote the resulting gameplay evidence by $e_k^0$ and use it to plan the next version. A Baseline Auditor checks the game and replay evidence. An Evolution Planner, acting as the Designer, then proposes 1 game-specific gameplay and aesthetic goal $g_k$ with up to 3 repair milestones:
\begin{equation}
\label{eq:rsi-goal}
e_k^0=\operatorname{Play}(P_k),\qquad
g_k=\operatorname{Plan}(x,P_k,e_k^0).
\end{equation}
The plan defines a core playtest and an upgrade playtest. The action interface and declared test entries remain fixed within each cycle, while gameplay and content can be extended through these existing interfaces.

An Execution Planner specifies the code and material changes for the complete upgrade. The Builder Agent implements them. A Goal Judge admits the build for repair when core gameplay is playable, evidence is valid, and no critical regression is observed. The admitted build becomes $P_{k,0}$. Otherwise, we set $P_{k,0}=P_k$ and run no repair rounds.

Each admitted repair round starts with fresh Player Agent evidence $e_{k,r}$ from the current project $P_{k,r}$. The Quality Judge scores gameplay design and aesthetics against the upgrade goal. A Diagnosis Agent identifies issues that limit these scores and attributes their causes. The Execution Planner proposes coordinated changes, which the Builder Agent implements:
\begin{equation}
\label{eq:rsi-candidate}
\begin{aligned}
q_{k,r}&=\pi_{\mathrm{plan}}(P_{k,r},g_k,e_{k,r}),\\
(\Delta C_{k,r},\Delta A_{k,r})
  &=\pi_{\mathrm{build}}(P_{k,r},q_{k,r}),\\
\widetilde P_{k,r}
  &=\operatorname{Apply}(P_{k,r},\Delta C_{k,r},\Delta A_{k,r}).
\end{aligned}
\end{equation}
Here $q_{k,r}$ is the Execution Planner's proposal. The material change $\Delta A_{k,r}$ may revise the video generation prompt, motion, generation scope, or runtime binding. The code change $\Delta C_{k,r}$ updates gameplay and composition. Fresh Player policies test both versions against the same core and upgrade goals. We denote the paired replay evidence from $P_{k,r}$ and $\widetilde P_{k,r}$ by $e'_{k,r}$. An A/B Judge retains the candidate only when valid replay evidence shows an improvement, no evaluated track regresses, and no critical regression occurs:
\begin{equation}
\label{eq:rsi-selection}
P_{k,r+1}=
\operatorname{Select}_{\mathrm{evolve}}
(P_{k,r},\widetilde P_{k,r},e'_{k,r}).
\end{equation}
After $R_k\leq 3$ executed repair rounds, the last admitted version becomes $P_{k+1}=P_{k,R_k}$. The recurrence can continue with $P_{k+1}$ as the input to the next evolution cycle.

\section{Experiments}
\label{sec:experiments}

\begin{figure}[htbp]
  \centering
  \captionsetup{skip=2pt}
  \includegraphics[width=\textwidth]{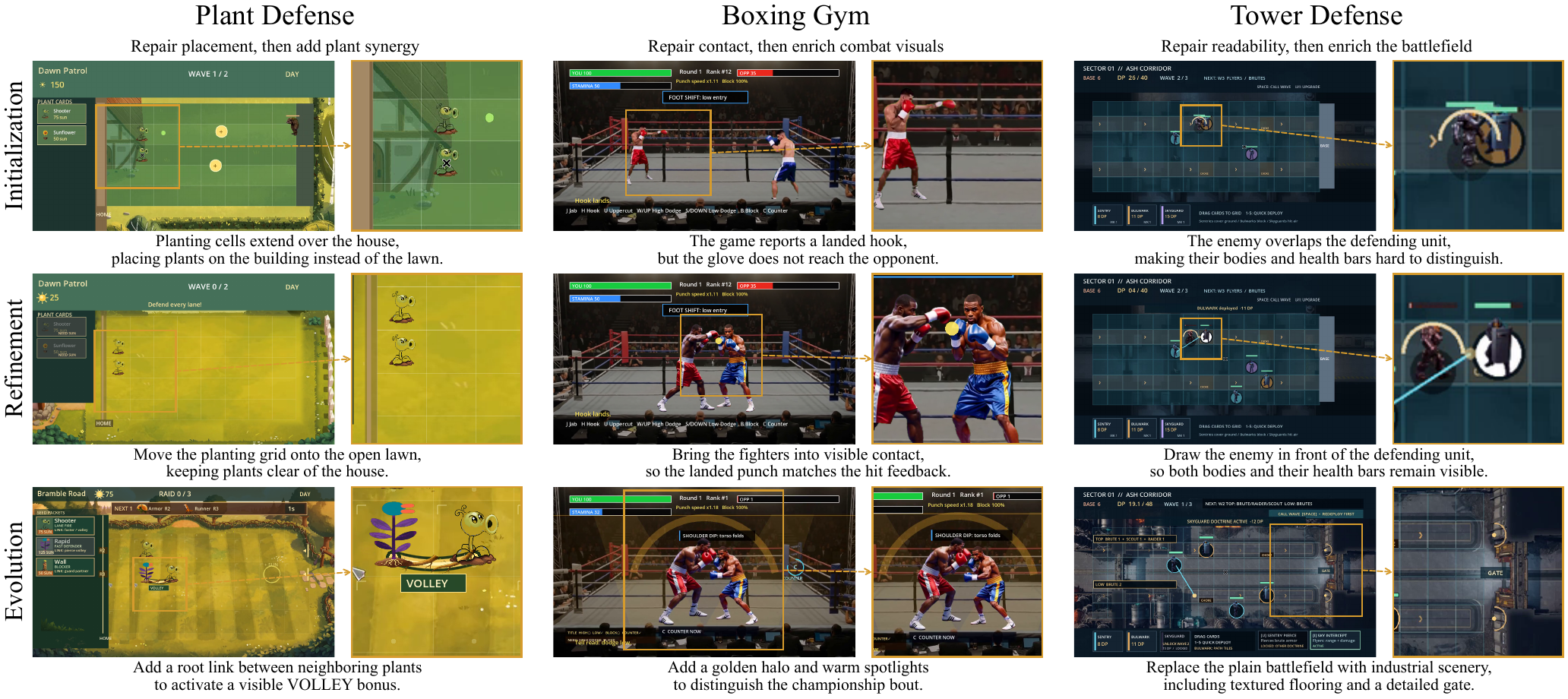}\par
  \caption{Gameplay at \textrm{V0}, \textrm{V1}, and \textrm{V2} for 3 games, with each pair showing the full frame (left) and an enlarged view of its yellow-boxed region (right). Game Refinement repairs visible defects, while Game Evolution enriches scene appearance and gameplay.}
  \label{fig:main-result-grid}
\end{figure}

\subsection{Implementation details}
Codex with GPT-6 Sol~\cite{openai2026gpt6sol} at medium reasoning effort handles construction, planning, and code editing. The Player and Judge also use GPT-6 Sol at medium reasoning effort. MiniMax H3~\cite{minimax2026h3} generates game materials on 8 NVIDIA H200 GPUs. We support Godot and browser-based HTML games, with code controlling input and state and generated materials supplying visuals. Replay adapters execute inputs and record gameplay. Game Refinement uses 3 planned playtest groups with up to 3 repair rounds per group. We record \textrm{V2} after 1 evolution cycle with up to 3 repair rounds. The successive-evolution experiment runs 3 consecutive cycles, though further cycles are supported. Skill proposals summarize at most 8 rounds of behavior results. The Quality Judge and A/B Judge assess current replay evidence without round history. Further implementation details appear in the supplementary material.

\subsection{Benchmark and evaluation protocol}
We evaluate frozen projects on GameCraft-Bench~\cite{luo2026gamecraft}, with 140 tasks across 15 categories, and GameASG-Bench~\cite{zhang2026gameasg}, with 47 tasks across 12 categories. GameCraft-Bench reports build rate and scores for mechanics (M), depth (D), functional visuals (V), art (A), and task-weighted Overall. GameASG-Bench reports static compliance (L1), runtime behavior (L2), and strict task success. L2 covers startup (P0), required gameplay (P1), and extended features (P2). Benchmark judgments are withheld from optimization. Within each benchmark, we compute means over the same game set for all compared methods and versions. Scoring details appear in the supplementary material.

\subsection{Main results}
We denote the games after Initialization, Refinement, and the first Evolution cycle by \textrm{V0}, \textrm{V1}, and \textrm{V2}, respectively. Further evolution cycles yield \textrm{V3}, \textrm{V4}, and so on. Our primary baseline is adapted Play2Code~\cite{huang2026play2code}. Its agent directly diagnoses gameplay issues and can revise code or call H3 to generate video materials. For a fair comparison, we use the same total development-token budget for Play2Code and our method. We discuss the concurrent works RSIGame~\cite{wu2026rsigame} and Recursive Game Creator (RGC)~\cite{chen2026recursive} further in the supplementary material.

\noindent\textbf{Quantitative results.} Tab.~\ref{tab:benchmark-main} and Tab.~\ref{tab:gameasg-main} report results on the full task sets. GameCraft-Bench Overall scores rise from 50.81 at \textrm{V0} to 54.45 at \textrm{V1} and 60.05 at \textrm{V2} (Tab.~\ref{tab:benchmark-main}). The \textrm{V2} mean exceeds Play2Code's 52.46. We use a dedicated trajectory comparison protocol to ensure fair evaluation across game versions. The supplementary material details this protocol. Across all evaluated GameASG-Bench games at each checkpoint, strict task success rises from 35.7\% at \textrm{V0} to 57.4\% at both \textrm{V1} and \textrm{V2}. Mean L2 rises from 89.3 to 94.5 and then stays at 94.5 (Tab.~\ref{tab:gameasg-main}). The benchmark emphasizes required functionality and does not separately score aesthetics. Visual and optional gameplay upgrades may therefore leave these metrics unchanged. We examine this interpretation in the supplementary material.

\begin{table}[htbp]
  \centering
  \footnotesize
  \setlength{\tabcolsep}{4pt}
  \renewcommand{\arraystretch}{1.08}
  \captionsetup{skip=3pt}
  \caption{GameCraft-Bench scores and development tokens, with token counts for our checkpoints reporting the additional cost of each stage. Upgraded games have the highest mean Overall score.}
  \label{tab:benchmark-main}
  \begin{tabular}{@{}lrrrrrr@{}}
    \hline
    Method & M & D & V & A & Overall & \shortstack{Tok.\\(M)} \\
    \hline
    Initialization~(\textrm{V0}) & 68.70 & 48.83 & 51.63 & 44.76 & 50.81 & 0.30 \\
    \hline
    Play2Code & 69.84 & 50.07 & 51.95 & 51.77 & 52.46 & 8.89 \\
    Refinement~(\textrm{V1}) & 72.97 & 52.38 & 52.50 & 49.35 & 54.45 & 3.69 \\
    Evolution~(\textrm{V2}) & \textbf{78.02} & \textbf{56.39} & \textbf{57.51} & \textbf{57.03} & \textbf{60.05} & 4.89 \\
    \hline
  \end{tabular}
  
\end{table}

\begin{table}[htbp]
  \centering
  \footnotesize
  \setlength{\tabcolsep}{4pt}
  \renewcommand{\arraystretch}{1.08}
  \captionsetup{skip=3pt}
  \caption{GameASG-Bench scores after Initialization, Refinement, and Evolution. Refinement and Evolution improve aggregate strict success and mean L2 over Initialization.}
  \label{tab:gameasg-main}
  \begin{tabular}{@{}lrrrrrr@{}}
    \hline
    Stage & \shortstack{Task\\success} & L1 & \multicolumn{4}{c}{L2} \\
    \cline{4-7}
     & & & Mean & P0 & P1 & P2 \\
    \hline
    Initialization~(\textrm{V0}) & 35.7\% & 92.7 & 89.3 & 96.7 & 88.7 & 88.7 \\
    Refinement~(\textrm{V1}) & \textbf{57.4\%} & \textbf{95.1} & \textbf{94.5} & \textbf{100.0} & 94.5 & \textbf{95.7} \\
    Evolution~(\textrm{V2}) & \textbf{57.4\%} & \textbf{95.1} & \textbf{94.5} & \textbf{100.0} & \textbf{94.7} & 95.4 \\
    \hline
  \end{tabular}
  \par
\end{table}

For human evaluation, we selected 30 games and asked 12 participants to rate anonymous Play2Code and \textrm{V0}/\textrm{V1}/\textrm{V2} clips. They scored visual appeal, motion quality, artifact absence, willingness to play, and overall quality on 5-point scales (Tab.~\ref{tab:human-stage-ratings}). We average ratings within each game before averaging across games. \textrm{V2} receives the highest mean on all 5 dimensions, including overall quality (3.67 versus 2.43 for Play2Code). The full protocol is in the supplementary material.

\begin{table}[htbp]
  \centering
  \small
  \setlength{\tabcolsep}{4pt}
  \renewcommand{\arraystretch}{1.08}
  \captionsetup{skip=3pt}
  \caption{Human ratings compare adapted Play2Code with the \textrm{V0}, \textrm{V1}, and \textrm{V2} checkpoints on a 5-point scale. \textrm{V2} receives the highest mean rating across all 5 dimensions.}
  \label{tab:human-stage-ratings}
  \begin{tabular}{@{}lc|ccc@{}}
    \hline
    Dimension (1--5) & \textrm{V0} & Play2Code & \textrm{V1} & \textrm{V2} \\
    \hline
    Visual appeal & 2.79 & 2.80 & 3.52 & \textbf{3.87} \\
    Motion quality & 2.41 & 2.53 & 3.10 & \textbf{3.67} \\
    Artifact absence & 2.56 & 2.77 & 3.29 & \textbf{3.52} \\
    Willingness to play & 2.32 & 2.33 & 3.10 & \textbf{3.56} \\
    Overall quality & 2.38 & 2.43 & 3.27 & \textbf{3.67} \\
    \hline
  \end{tabular}
  
\end{table}

\noindent\textbf{Qualitative results.} Fig.~\ref{fig:main-result-grid} shows Plant Defense, Boxing Gym, and Tower Defense, with a gameplay frame and an enlarged detail at each checkpoint. Refinement corrects planting placement, punch contact, and overlapping combat units. Evolution adds linked-plant bonuses, championship lighting, and a detailed industrial battlefield. Further examples, including browser games, appear in the supplementary material.

\subsection{Successive game evolution}
\label{sec:successive-upgrades}
The preceding results evaluate 1 evolution cycle. Here we examine whether games continue to improve across successive evolution cycles. We selected 10 games and ran 3 consecutive evolution cycles, each starting from the previously retained version. Every cycle proposes a new objective and allows up to 3 candidate rounds. Tab.~\ref{tab:upgrade-trajectory} reports mean GameCraft-Bench scores from \textrm{V0} through \textrm{V4}.

\begin{table}[htbp]
  \centering
  \small
  \setlength{\tabcolsep}{5pt}
  \caption{GameCraft-Bench means from a 10-game successive-evolution study. Overall rises across successive checkpoints.}
  \label{tab:upgrade-trajectory}
  \begin{tabular}{@{}lrrrrr@{}}
    \hline
    Version & M & D & V & A & Overall \\
    \hline
    \textrm{V0} & 66.25 & 53.46 & 61.80 & 42.33 & 52.74 \\
    \textrm{V1} & 73.80 & 57.18 & 65.34 & 44.87 & 56.59 \\
    \textrm{V2} & 74.05 & 57.32 & 68.57 & 47.88 & 58.21 \\
    \textrm{V3} & 74.15 & 58.21 & 69.07 & 51.59 & 59.91 \\
    \textrm{V4} & \textbf{74.30} & \textbf{63.57} & \textbf{72.96} & \textbf{53.29} & \textbf{62.99} \\
    \hline
  \end{tabular}
  \par
\end{table}

\begin{figure}[htbp]
  \centering
  \captionsetup{skip=2pt}
  \includegraphics[width=\linewidth]{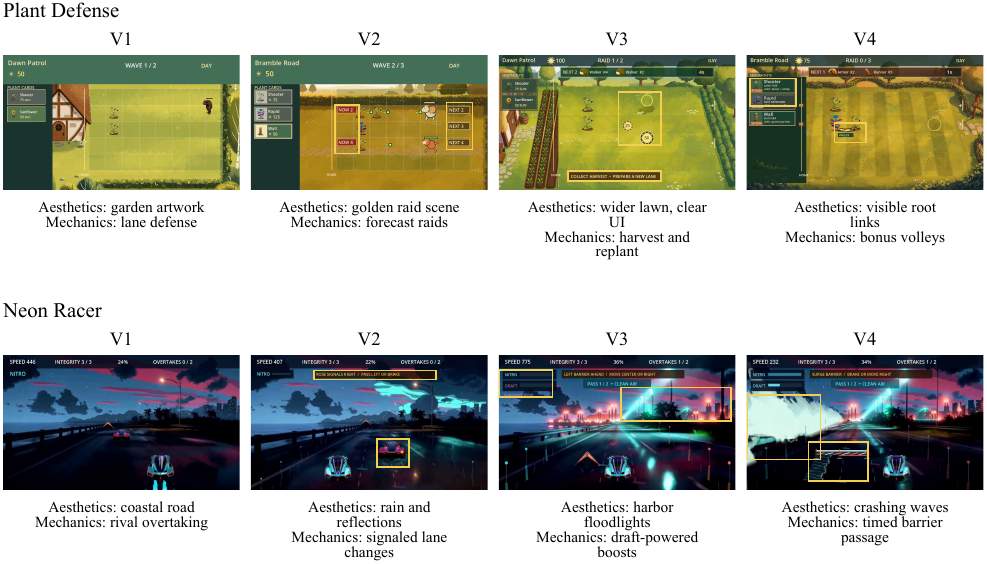}\par
  \caption{Successive evolution of two Godot games from \textrm{V1} to \textrm{V4}. The highlighted regions show how each cycle adds player choices and visual feedback.}
  \label{fig:upgrade-demos}
  
\end{figure}

Fig.~\ref{fig:upgrade-demos} traces 2 examples through 4 versions. Plant Defense adds 1) threatened-lane forecasts and mixed raids, 2) a harvest and new placement between raids, and 3) bonuses for adjacent plants. Neon Racer adds 1) signaled lane changes, 2) draft-powered boosts and harbor lighting, and 3) storm surges with timed road barriers. The screenshots use version-specific play routes, so they are not matched-input comparisons. We also count upgrades that are accepted, build successfully, and meet their stated goals. Internal acceptance alone does not establish a benchmark gain.

\begin{figure}[htbp]
  \centering
  \captionsetup{skip=2pt}
  \includegraphics[width=\textwidth]{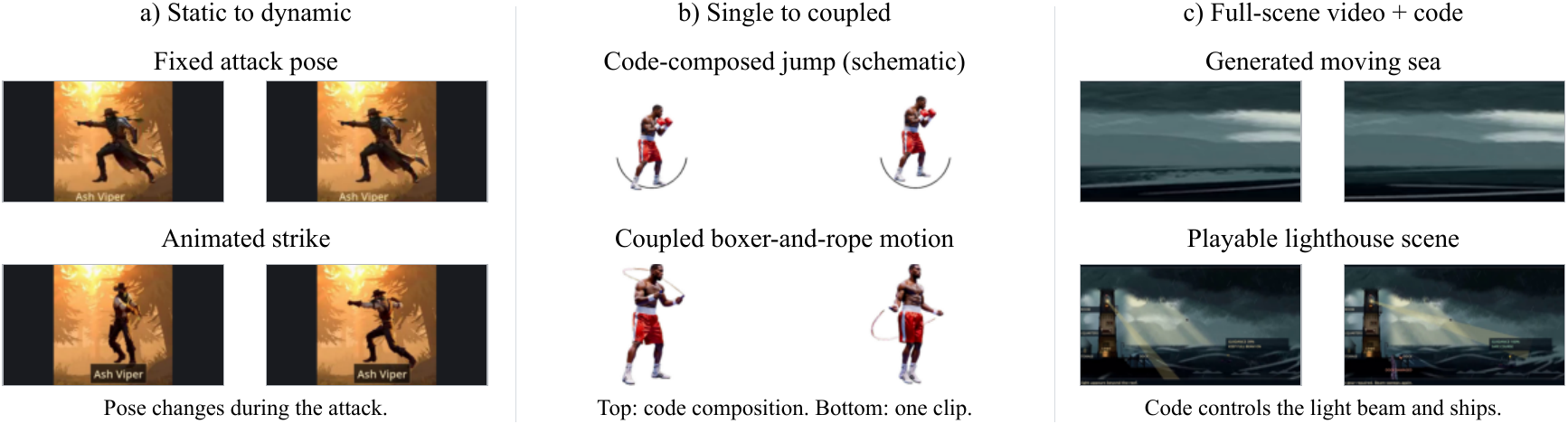}
  \caption{Three cases show how our agents use video generation to manage game assets and represent complex motion and object interactions. Panels (a), (b), and (c) show static-to-dynamic animation, coupled boxer-and-rope motion, and full-scene video integrated with game code.}
  \label{fig:material-management-cases}
  
\end{figure}

\begin{figure}[htbp]
  \centering
  \captionsetup{skip=2pt}
  \includegraphics[width=\linewidth]{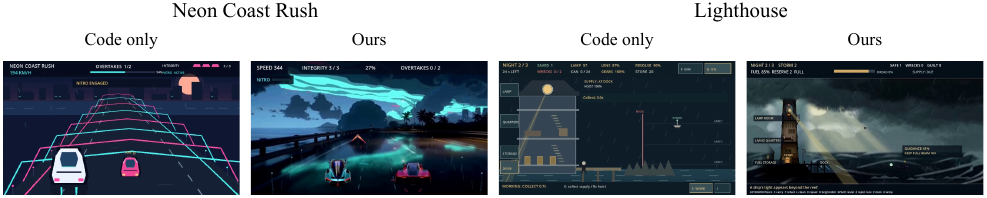}
  \caption{Code-only and video-assisted gameplay in Neon Coast Rush and Lighthouse. Video materials enrich roadside scenery, road lighting, and the stormy sea in these examples.}
  \label{fig:video-vs-code}
  
\end{figure}

\subsection{Adaptive video material management}
\noindent\textbf{Video-generated materials enrich game appearance.} Fig.~\ref{fig:video-vs-code} shows richer lighting in Neon Coast Rush and a stormy sea in Lighthouse compared with code-only implementations. These qualitative examples use different generation pipelines.

\noindent\textbf{Different generation levels support actor motion, coupled interactions, and scene dynamics.} Fig.~\ref{fig:material-management-cases} illustrates changes at the actor, interaction, and scene levels. In (a), Bounty Hunter replaces a fixed attack pose with articulated motion. In (b), the upper schematic illustrates a separately generated boxer and code-drawn rope. The selected coupled clip instead generates the boxer and moving rope together. In (c), Lighthouse uses a full-scene clip for the looping storm sea, while code controls lighting and ships. These cases show how generation scope assigns motion to video while retaining gameplay interactions in code.

\noindent\textbf{Recursive development expands both static and dynamic material inventories.} Tab.~\ref{tab:video-material-stages} shows growth in both static and dynamic materials. Between \textrm{V0} and \textrm{V2}, our method increases static materials from 120 to 257 and dynamic materials from 73 to 122. Play2Code adds 6 static materials and no dynamic materials. This expansion supplies additional appearance and motion resources for game development, although inventory counts alone do not establish quality gains.

\begin{table}[htbp]
  \centering
  \small
  \captionsetup{skip=3pt}
  \setlength{\tabcolsep}{2pt}
  \caption{Material inventories across our checkpoints and Play2Code. Dynamic materials increase from 73 at \textrm{V0} to 122 at \textrm{V2}, while Play2Code adds only 6 static materials.}
  \label{tab:video-material-stages}
  \begin{tabular}{@{}lr|rrr@{}}
    \hline
    Material level & \textrm{V0} & Play2Code & \textrm{V1} & \textrm{V2} \\
    \hline
    Static & 120 & 126 & 175 & 257 \\
    Single dynamic & 55 & 55 & 73 & 89 \\
    Coupled dynamic & 12 & 12 & 18 & 19 \\
    Full-scene dynamic & 6 & 6 & 6 & 14 \\
    \hline
    Dynamic total & 73 & 73 & 97 & 122 \\
    All materials & 193 & 199 & 272 & 379 \\
    \hline
  \end{tabular}
  \par
\end{table}

\subsection{Playtest Policies and Behavior Skill}
\label{sec:behavior-skill-effect}
We compare package traces, direct Codex and Qwen3.8-27B~\cite{qwen2026qwen38} Players, and Codex-authored play programs using functional coverage, exploration breadth and depth, and resource use. Breadth tracks gameplay branches, while depth tracks milestone progress~\cite{ferdous2021search,du2023humanexploration,wang2024voyager}. Detailed metric definitions are provided in the supplementary material.

We test playtest policies on 10 frozen games, with 20 decisions and 30 frames per decision for each scenario. Tab.~\ref{tab:player-comparison} reports mean results and supports three insights.

\noindent\textbf{1) Direct Codex achieves the strongest coverage, at a high cost.} It reaches 83.9\% functional coverage, 82.1\% breadth, and 76.2\% depth, matching or exceeding every other policy. Direct interaction allows Codex to adapt its decisions to observed gameplay, but requires 16.1 minutes and 1,001.9K tokens per game.

\noindent\textbf{2) Program-based play is more efficient, and Behavior Skill improves exploration with modest additional cost.} Codex-authored programs reduce time to 10.2 minutes and tokens to 201.3K. Adding Behavior Skill raises breadth from 73.0\% to 78.1\% and depth from 69.3\% to 75.9\%, while functional coverage reaches 83.9\%. This adds 1.4 minutes and 33.5K tokens per game, keeping the total at 11.6 minutes and 234.8K tokens.

\noindent\textbf{3) Qwen provides useful coverage with fewer tokens.} It reaches 76.8\% functional coverage and 74.9\% depth, below Direct Codex, while using 277.0K versus 1,001.9K tokens per game.

\begin{table}[htbp]
  \centering
  \small
  \setlength{\tabcolsep}{2pt}
  \caption{Playtest coverage and resource use across 10 games. All reported means are computed over the same 10 games for every policy.}
  \label{tab:player-comparison}
  \begin{tabular}{@{}lrrrrr@{}}
    \hline
    Policy & Func. & Breadth & Depth & Min. & Tok. (K) \\
    \hline
    \textit{Package traces} & 82.1\% & 71.0\% & 63.8\% & \textbf{6.9} & 23.6 \\
    \textit{Direct Codex} & \textbf{83.9\%} & \textbf{82.1\%} & \textbf{76.2\%} & 16.1 & 1,001.9 \\
    \textit{Direct Qwen 27B} & 76.8\% & 70.2\% & 74.9\% & 16.6 & 277.0 \\
    \hline
    \textit{Codex program} & 80.4\% & 73.0\% & 69.3\% & 10.2 & 201.3 \\
    \quad + Behavior Skill & \textbf{83.9\%} & 78.1\% & 75.9\% & 11.6 & 234.8 \\
    \hline
  \end{tabular}
\end{table}

Further discussion and results are provided in the supplementary material.

\section{Conclusion}
\label{sec:conclusion}

We present Play2Evolve, a recursive game development framework that uses agent playtest feedback to repair defects and guide successive upgrades. Video generation and code jointly enrich visual assets while preserving interactive control. Benchmark and human evaluations show gains over initialization and Play2Code, with further analyses supporting adaptive material management and efficient exploration through Behavior Skill.

\clearpage
{\small
\bibliographystyle{ieeenat_fullname}
\bibliography{main}}
\clearpage
\appendix
\makeatletter
\setlength{\@fptop}{0pt}
\makeatother
\section*{Supplementary Material}
This supplementary material details our evaluation protocols, provides additional game examples, and examines benchmark limitations. The sections are organized as follows.

\begin{itemize}
  \item Sec.~\ref{sec:supp-final-showcase} describes how demonstrations are prepared, verified, and packaged after Game Refinement.
  \item Sec.~\ref{sec:supp-rsi-showcase} explains how trajectories adapt to revised games while preserving shared demonstration targets and evaluation budgets.
  \item Sec.~\ref{sec:supp-playtest-metrics} defines functional coverage, exploration breadth, and exploration depth through a worked game example.
  \item Sec.~\ref{sec:supp-playtest-policy-comparison} details the Player comparison, resource accounting, and paired observations with Behavior Skill.
  \item Sec.~\ref{sec:supp-successive-upgrades} presents successive-evolution results, function addition and loss metrics, and a comparison protocol for continued Refinement versus Evolution.
  \item Sec.~\ref{sec:supp-astra-three-round} compares Sol and Astra backbones across Initialization, Refinement, and Evolution.
  \item Sec.~\ref{sec:supp-rsigame-rgc} compares scores and gameplay examples with concurrent work, RSIGame and RGC, on 10 selected games.
  \item Sec.~\ref{sec:supp-stage-visualizations} provides additional gameplay examples across Initialization, Refinement, and Evolution.
  \item Sec.~\ref{sec:scoring} analyzes how demonstration selection, visual sampling, and functional checks affect GameCraft-Bench and GameASG-Bench scores.
  \item Sec.~\ref{sec:supp-main-protocol} specifies the main comparison setup, benchmark aggregation, and development-token accounting.
  \item Sec.~\ref{sec:supp-behavior-skill} describes the Behavior Skill evaluation and illustrates saved guidance with historical playtest examples.
  \item Sec.~\ref{sec:supp-human-stages} describes the human-evaluation questionnaire, rating criteria, and aggregation procedure.
  \item Sec.~\ref{sec:discussion} summarizes limitations in exploration, execution evidence, candidate improvement, and evaluation scope.
\end{itemize}

\section{Final Showcase and Demonstration Packaging}
\label{sec:supp-final-showcase}
After program-based refinement, a separate demonstration-preparation stage collects replay evidence for the selected final game. It does not edit game code or materials, and it does not compute a benchmark score. The stage can be disabled when the earlier declared-trace packaging protocol is required.

\begin{enumerate}
  \item \textbf{Freeze the selected game.} The stage reads the selected project, its Behavior Skill when available, the Initialization recordings, and the final game's recordings for the original scene groups. Newly authored policies may use the Behavior Skill. The selected project is checked for changes before packaging.

  \item \textbf{Preserve the Initialization scene coverage.} For each original scene group, the stage compares visible facts from its Initialization recording with an official replay of the final game's group trace. Missing facts trigger a new Player policy for the final game. At most 2 revisions follow the first attempt. Several selected clips may jointly preserve 1 group's visible facts. If the Initialization coverage floor remains unmet, the showcase cannot publish the package.

  \item \textbf{Record uncovered public requirements.} A visual assessment labels each requirement as shown, partial, not shown, or unverified using frame evidence. Only requirements not shown by the selected group clips are sent to the extra-clip planner. Each extra profile names an observable goal, an existing launch, relevant requirements, and a frame budget of at most 600. A missed extra profile permits 1 targeted revision. Unassigned requirements remain explicit in the manifest. The total budget is ten selected clips, including the original group clips.

  \item \textbf{Verify and package the traces.} The selected actions are replayed through the official OS-input recorder on the frozen final project. The exported action traces must match the inputs used for those replays. The final package includes the selected policies, replay videos, file hashes, and coverage reports. Replay verification establishes that a video came from its recorded actions on the selected game. It does not establish complete requirement coverage, a clean benchmark build, or a benchmark score.
\end{enumerate}

\section{Trajectory Module for Fair GameCraft-Bench Evaluation}
\label{sec:supp-rsi-showcase}
GameCraft-Bench judges games through recorded gameplay. A feature can be implemented correctly yet remain unseen when a trajectory fails to reach it. Changes in timing, controls, or navigation can also make an earlier input sequence unsuitable for an evolved game. Our trajectory module prepares replayable demonstrations on frozen projects, preserving shared interaction targets while adapting inputs to each version. GameASG-Bench instead executes its own functional tests and does not use this demonstration-preparation procedure.

\paragraph{Shared interaction targets.}
The module first replays each reference trajectory on both the reference and evaluated projects. A visual comparison extracts observable facts from the reference recording, including game modes, interactions, transitions, and major outcomes. These facts define what the corresponding demonstration should preserve. For example, a route showing a purchase followed by combat should continue to expose both interactions after evolution. Matching these events does not require identical transient counters or animation frames. We retain the same launch scenario and 1 demonstration slot per reference trajectory, with at most 10 slots.

\paragraph{Bounded trajectory adaptation.}
The original input sequence remains a candidate. The module also considers compatible playtest traces recorded on the exact evaluated project and replays them through the official recorder. Candidates are ranked first by the number of reference facts they preserve. If facts remain missing, the Player can generate at most 2 replacement programs for that slot, each limited to 600 simulation frames. The Player receives the reference inputs and missing facts so that it can adjust the route to the game's observed behavior. Trajectory preparation changes neither game code nor visual materials.

\paragraph{Showing newly introduced gameplay.}
Evolution can introduce interactions absent from the reference game. Among candidates preserving the same number of reference facts, the module favors those showing more of the declared evolution goals. This allows new content to appear within the existing demonstration slots while giving priority to shared interactions. Goal visibility is assessed from replay evidence and is separate from the benchmark score. New goals do not trigger additional repair attempts once the reference facts are preserved.

\paragraph{Independent scoring and comparison conditions.}
Selected traces are replayed on the frozen project and checked again against the reference facts before benchmark evaluation. Project and trace hashes link the submitted recordings to the evaluated checkpoint. Benchmark scores are not used to select or revise trajectories. Across compared methods, we use the same rules for constructing reference routes, the same Player configuration, and the same limits on demonstrations and trajectory repair. This protocol aligns the opportunity to demonstrate game behavior while allowing version-specific input sequences. Any reference facts that remain unobserved are recorded explicitly, since a successful replay alone does not establish complete interaction coverage.

\section{Playtest Coverage Metrics}
\label{sec:supp-playtest-metrics}
We use 3 complementary metrics to audit what a playtest reaches. The Sokoban Dungeon example below has 3 named scenarios. Each policy receives the same scenario entry and decision budget. We compute the metrics from recorded game states after play, not from actions attempted by the policy. This is a retrospective pilot audit rather than a preregistered evaluation.

\begin{enumerate}
  \item \textbf{Functional coverage.} Count target gameplay events that actually occur across the 3 scenarios, then divide by the 12 listed events. An attempted action alone does not count. The 4 targets for each scenario are:
  \begin{itemize}
    \item Title-to-dungeon flow: enter the dungeon, push a crate, open the pressure-plate gate, and descend to floor 2.
    \item Heavy-crate puzzle: brace the heavy crate, push it, collect the blue key, and open the blue gate.
    \item Freeze-item puzzle: use the freeze ability, push a crate, collect the chest item, and descend to floor 5.
  \end{itemize}

  \item \textbf{Exploration breadth.} Count distinct gameplay branches newly entered from a scenario's initial state, then divide by 7. The branch catalog is the title screen, the death state, and floors 1 through 5. A branch present at the scenario entry does not count unless the player leaves and enters it again. Repeated entries into the same branch count once across all 3 scenarios. The numerator counts observed state transitions, not distinct positions within a floor.

  \item \textbf{Exploration depth.} Each scenario has 4 ordered milestones. Its score is the index of the furthest milestone reached, from 0 to 4, divided by 4. The game-level score is the mean over the 3 scenarios. Earlier milestones need not be observed when a later milestone is reached. The milestone lists are:
  \begin{itemize}
    \item Title-to-dungeon flow: enter the dungeon, push a crate, open the pressure-plate gate, and reach floor 2.
    \item Heavy-crate puzzle: brace the heavy crate, push it, collect the blue key, and reach floor 4.
    \item Freeze-item puzzle: use the freeze ability, collect the chest item, reach floor 5, and reach floor 6. Reaching floor 5 as the furthest milestone scores 3/4 for this scenario.
  \end{itemize}
\end{enumerate}

These metrics capture different aspects of play. Functional coverage checks specified events, breadth checks newly visited branches, and depth checks progression within a scenario. The distinct-position count in an earlier 2-game pilot is a separate exploratory measure and is not the exploration-breadth numerator defined here.

\section{Playtest Policy Comparison}
\label{sec:supp-playtest-policy-comparison}
We compare package traces, direct model-based Players, and Codex-authored play programs on the same frozen games. All policies receive the same scenario entries, goals, and action budgets. Package traces replay the actions provided with each scenario, with truncation or waiting used to match the action budget. These traces target their scenario goals and serve as an informed reference. Policy results are aggregated over the same game set.

\paragraph{Coverage and exploration.}
We compute functional coverage, exploration breadth, and exploration depth from recorded game states using the definitions in Sec.~\ref{sec:supp-playtest-metrics}. Functional coverage and breadth use the union of observed events and branches across scenarios. Depth averages the normalized furthest milestone reached within each scenario. We first aggregate within each game and then average games equally. Branches and progression milestones use separate catalogs, so reaching a later milestone need not increase breadth.

\paragraph{Behavior Skill comparison.}
The Skill-conditioned program generator receives guidance saved from earlier playtests. The comparison program receives no Skill input. Both conditions use the same frozen project, scenario goals, Player configuration, and action budget. The Skill is fixed before the evaluated playtest. The comparison measures differences in coverage, exploration, and resource use, with aggregate results reported in the main paper.

\paragraph{Resource accounting.}
Playtest time includes policy generation where applicable, game execution, and replay verification. Recorded tokens include input and output tokens, including cached input. Game development, independent benchmark judging, video generation, and earlier Skill creation are excluded from the Player comparison. Package traces require no model calls during replay, so their preparation-token estimate comes from a separate trace-authoring call using game instructions and scenario goals. This proxy is distinct from the measured cost of producing and validating the original traces. Tokenizers and prompt formats differ across models, so token totals describe the tested configurations rather than normalized compute costs.

\section{Successive Evolution Study Protocol}
\label{sec:supp-successive-upgrades}
We evaluate 10 games over 3 consecutive Evolution cycles, starting from the refined \textrm{V1} checkpoint. This fixed 10-game subset is distinct from the full-cohort results reported in the main paper; its \textrm{V1}--\textrm{V4} scores are therefore compared only within this study. Each cycle starts from the previously retained project and allows up to 3 candidate rounds. The selected outputs are denoted by \textrm{V2}, \textrm{V3}, and \textrm{V4}. The mean GameCraft-Bench Overall scores match the main paper, rising from 56.59 at \textrm{V1} to 58.21, 59.91, and 62.99. Evaluation follows the trajectory protocol in Sec.~\ref{sec:supp-rsi-showcase}.

\subsection{New Functionality and Loss of Existing Functions}
\label{sec:supp-upgrade-function-audit}
Beyond benchmark scores, we examine how much functionality each cycle adds and how much existing functionality it loses. Each function is an observable gameplay capability with a consistent identity and pass condition across checkpoints. New functions introduce interactions or gameplay options absent before the cycle. Repairing an existing capability or changing only its appearance does not count as adding a function. Restoring a previously lost function is recorded as a repair.

\paragraph{Function addition rate.}
For each game and cycle, we divide the number of verified newly introduced functions by the number of working functions in the preceding retained checkpoint. This measures the expansion relative to the existing game, rather than the fraction of proposed upgrade goals that were implemented. Each distinct new capability is counted once.

\paragraph{Existing-function loss rate.}
Using the same preceding-checkpoint checklist and denominator, we count previously working functions that are confirmed to be unavailable or broken after evolution. A function that is merely absent from a recorded route is marked unverified rather than counted as lost. Verification uses recorded interactions and state changes, independently of the internal candidate-selection decision. The checklist preserves function identities across cycles so that additions, regressions, and repairs can be distinguished.

Both rates are computed within each game before averaging across the same 10 games. If a candidate is rejected, we evaluate the retained checkpoint. Retaining an unchanged project introduces no new functions and causes no new functional loss. Tab.~\ref{tab:supp-upgrade-function-audit} pairs the function statistics with the corresponding benchmark scores.

\begin{table}[H]
  \centering
  \small
  \setlength{\tabcolsep}{3pt}
  \caption{Successive Evolution across 10 games, with Overall scores matching the main paper. Function addition and loss rates measure expansion and regression relative to the preceding checkpoint.}
  \label{tab:supp-upgrade-function-audit}
  \begin{tabular}{lrrr}
    \hline
    Checkpoint & Overall & \shortstack{Functions\\added (\%)} & \shortstack{Existing functions\\lost (\%)} \\
    \hline
    \textrm{V1} & 56.59 & -- & -- \\
    \textrm{V2} & 58.21 & 17 & 0 \\
    \textrm{V3} & 59.91 & 14 & 1 \\
    \textrm{V4} & 62.99 & 12 & 1 \\
    \hline
  \end{tabular}
\end{table}

\begin{figure}[!t]
  \centering
  \includegraphics[width=\textwidth]{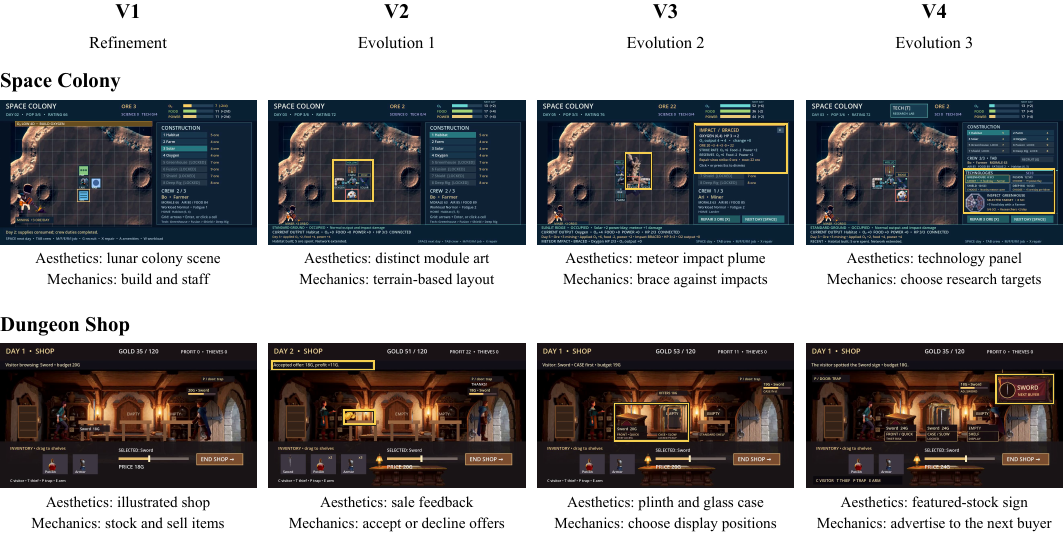}
  \caption{Successive Evolution from \textrm{V1} to \textrm{V4} in Space Colony and Dungeon Shop, with aesthetic and mechanic changes labeled below each frame. Yellow boxes locate visible changes as the games add terrain planning, meteor response, and research choices, or bargaining, display placement, and advertising.}
  \label{fig:supp-upgrade-trajectories}
\end{figure}

\paragraph{Qualitative examples.}
The main paper illustrates successive Evolution with Plant Defense and Neon Racer. Fig.~\ref{fig:supp-upgrade-trajectories} adds Space Colony and Dungeon Shop. Space Colony introduces terrain-aware placement, meteor response, and a research choice. Dungeon Shop adds bargaining, display placement, and featured-stock advertising. Each frame comes from a playtest of the corresponding retained checkpoint.

\subsection{Continued Refinement versus Evolution}
\label{sec:supp-repair-vs-upgrade}
This comparison tests whether explicit evolution planning adds value beyond continuing refinement. The comparison uses 15 games, with both branches starting from the same frozen \textrm{V1} checkpoint for each game. One branch continues Refinement against the original requirements. The other performs an Evolution cycle that proposes new gameplay and aesthetic goals. Both branches use the same agent backbone, Player, video generator, and starting Behavior Skill, with equal additional development-token and video-generation budgets. Candidate-evaluation and playtest budgets are also matched, and planning costs count toward the development budget.

Tab.~\ref{tab:supp-repair-vs-upgrade} reports GameCraft-Bench mechanics (M), depth (D), functional visuals (V), art (A), and Overall scores on the 15 games. The \textrm{V0} and \textrm{V1} rows provide reference checkpoints, and \textrm{V2} is the result of Evolution from \textrm{V1}. Both branches use the same independent scoring and trajectory protocol. The continued-Refinement Overall score uses weights of 0.15, 0.35, 0.15, and 0.35 for M, D, V, and A, respectively.

\begin{table}[H]
  \centering
  \small
  \setlength{\tabcolsep}{2pt}
  \caption{GameCraft-Bench scores for continued Refinement and Evolution on 15 games under equal additional budgets. Both branches start from the same \textrm{V1} checkpoints.}
  \label{tab:supp-repair-vs-upgrade}
  \begin{tabular}{lrrrrr}
    \hline
    Condition & M & D & V & A & Overall \\
    \hline
    \textrm{V0} & 66.25 & 53.46 & 61.80 & 42.33 & 52.74 \\
    \textrm{V1} & 73.80 & 57.18 & 65.34 & 44.87 & 56.59 \\
    \hline
    \shortstack[l]{\textrm{V1} +\\Refinement} & 73.75 & 57.24 & 65.20 & 45.02 & 56.63 \\
    \shortstack[l]{\textrm{V1} +\\Evolution (\textrm{V2})} & 74.05 & 57.32 & 68.57 & 47.88 & 58.21 \\
    \hline
  \end{tabular}

\end{table}

\section{Comparison across Agent Backbones}
\label{sec:supp-astra-three-round}

We evaluate Play2Evolve with Sol and Astra on 10 selected games using the same GameCraft-Bench protocol. Tab.~\ref{tab:supp-astra-three-round} compares Initialization (\textrm{V0}), Refinement (\textrm{V1}), and Evolution (\textrm{V2}) for both backbones. Both configurations improve across these checkpoints, with Overall rising from 50.81 to 60.98 for Sol and from 54.40 to 62.45 for Astra.

\begin{table}[H]
  \centering
  \small
  \setlength{\tabcolsep}{6pt}
  \caption{GameCraft-Bench Overall scores for Sol and Astra at 3 checkpoints on 10 selected games. Both backbones improve from Initialization to Evolution.}
  \label{tab:supp-astra-three-round}
  \begin{tabular}{lrrr}
    \hline
    Configuration & \textrm{V0} & \textrm{V1} & \textrm{V2} \\
    \hline
    Ours (Sol) & 50.81 & 54.67 & 60.98 \\
    Ours (Astra) & 54.40 & 58.52 & 62.45 \\
    \hline
  \end{tabular}
\end{table}

\section{Comparison with Concurrent Methods}
\label{sec:supp-rsigame-rgc}

RSIGame~\cite{wu2026rsigame} and Recursive Game Creator (RGC)~\cite{chen2026recursive} are concurrent work on agentic game development. We compare scores on 10 selected GameCraft-Bench games and examine representative gameplay outputs.

\paragraph{Experimental setup.}
We evaluated RSIGame and RGC on the selected 10 Godot games. For each task, both methods started from the same project and used GPT-6 Sol at medium reasoning effort. Each output was evaluated in a single GameCraft-Bench pass. Tab.~\ref{tab:supp-concurrent-scores} summarizes their mean Overall scores alongside our recorded Evolution (\textrm{V2}) scores for the same games. These results compare the evaluated configurations, which use different development procedures.

\begin{table}[H]
  \centering
  \small
  \setlength{\tabcolsep}{4pt}
  \caption{Mean GameCraft-Bench Overall scores on 10 selected games. Play2Evolve has the highest recorded mean among the evaluated configurations.}
  \label{tab:supp-concurrent-scores}
  \begin{tabular}{lrrr}
    \hline
    Metric & RSIGame & RGC & Ours (\textrm{V2}) \\
    \hline
    Overall & 48.29 & 33.29 & \textbf{62.42} \\
    \hline
  \end{tabular}
\end{table}

\FloatBarrier
\begin{figure}[!t]
    \centering
    \includegraphics[width=\textwidth]{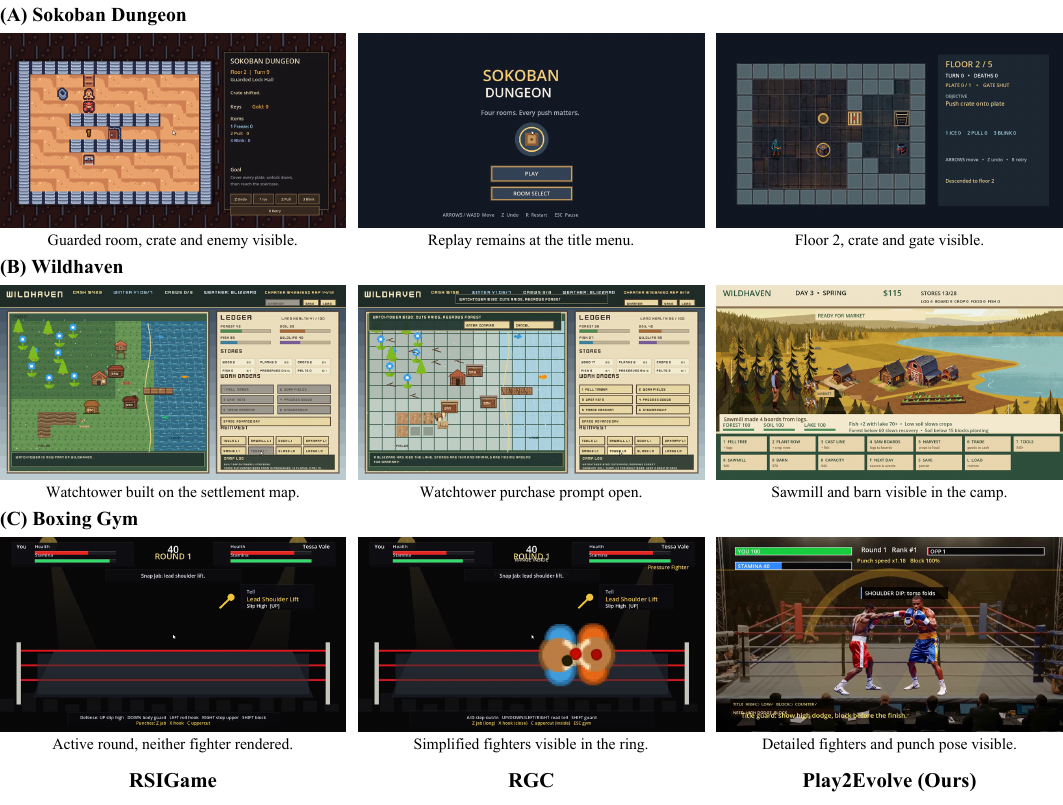}
    \caption{Qualitative gameplay comparison of RSIGame~\cite{wu2026rsigame}, Recursive Game Creator (RGC)~\cite{chen2026recursive}, and Play2Evolve on 3 shared tasks. The frames illustrate scenario access, visible buildings, and fighter rendering, with Play2Evolve shown from archived Evolution playtests using different trajectories.}
    \label{fig:supp-rsigame-rgc-gameplay}
\end{figure}

\paragraph{Qualitative comparison with Play2Evolve.}
Fig.~\ref{fig:supp-rsigame-rgc-gameplay} adds our archived Evolution outputs on 3 of these tasks. The RSIGame and RGC frames retain the original replay timestamps, 8 seconds for Sokoban Dungeon and Wildhaven and 4 seconds for Boxing Gym. Our frames come from task-specific playtests with different routes and durations. They illustrate the resulting gameplay and visual presentation, rather than equal-time progress under a shared action sequence.

\paragraph{Sokoban Dungeon.}
RSIGame reaches a guarded room with a crate and an enemy, while the displayed RGC replay remains at the title menu. The RGC run returned no selected project, so its integrated candidate was evaluated. Our frame shows floor 2 with a crate, a pressure plate, and a gate. These frames illustrate why entering a gameplay scene is necessary before judging its interactions. The floor number alone does not establish greater exploration depth across different game layouts.

\paragraph{Wildhaven.}
RSIGame shows a purchased watchtower on the settlement map, while RGC shows an open watchtower purchase prompt during a blizzard. Our development replay shows a sawmill and barn within a landscape containing forest, farmland, and a lake. Its resource display and activity message accompany the visible buildings. The examples illustrate different representations of settlement development, with our frame emphasizing the integration of detailed scenery, buildings, and gameplay state.

\paragraph{Boxing Gym.}
The RSIGame frame shows an active timer and health bars, but neither fighter is rendered. RGC shows simplified fighters at close range, while our frame shows detailed fighters in a punching pose against a populated arena. This makes character appearance and the combat pose directly visible. RGC scores 32.75 Overall compared with 29.56 for RSIGame, consistent with the rendering issue observed in this RSIGame case.

\paragraph{Discussion.}
The cases highlight scene access, observable interaction states, and visual presentation as distinct aspects of the resulting games. Play2Evolve's frames provide concrete examples of gameplay scenes with integrated visual assets. These selected frames support qualitative comparison, while claims about overall performance require the corresponding evaluation results.

\FloatBarrier
\section{Additional Game Examples across Checkpoints}
\label{sec:supp-stage-visualizations}
\begin{figure}[!t]
  \centering
  \captionsetup{skip=2pt}
  \includegraphics[width=0.85\textwidth]{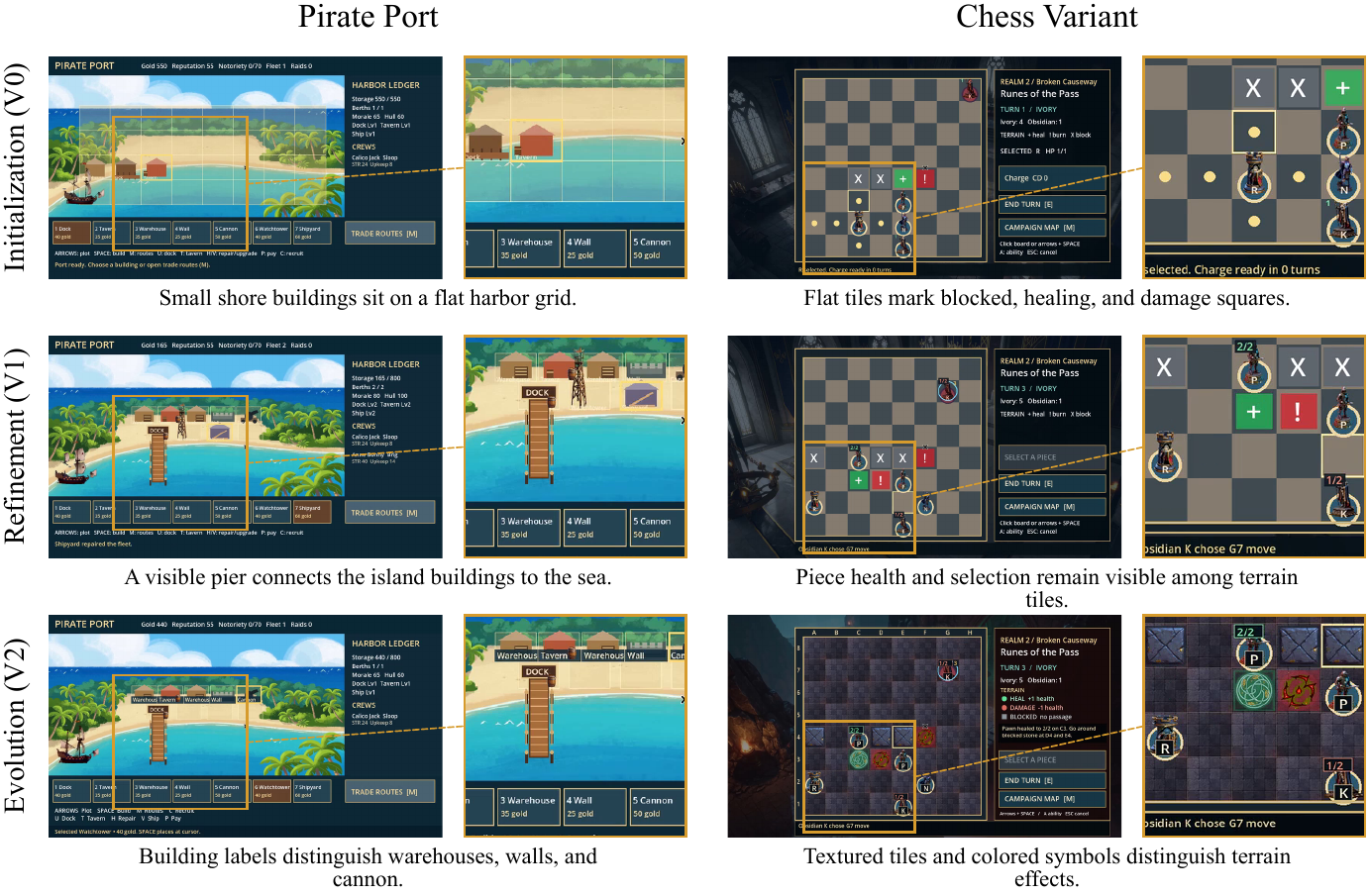}
  \caption{Pirate Port and Chess Variant across \textrm{V0}--\textrm{V2}, with a full frame and its enlarged detail at each checkpoint. The highlighted regions show changes to harbor structures, building labels, and terrain rendering.}
  \label{fig:supp-stage-godot-a}
\end{figure}

\begin{figure}[htbp]
  \centering
  \captionsetup{skip=2pt}
  \includegraphics[width=0.95\textwidth]{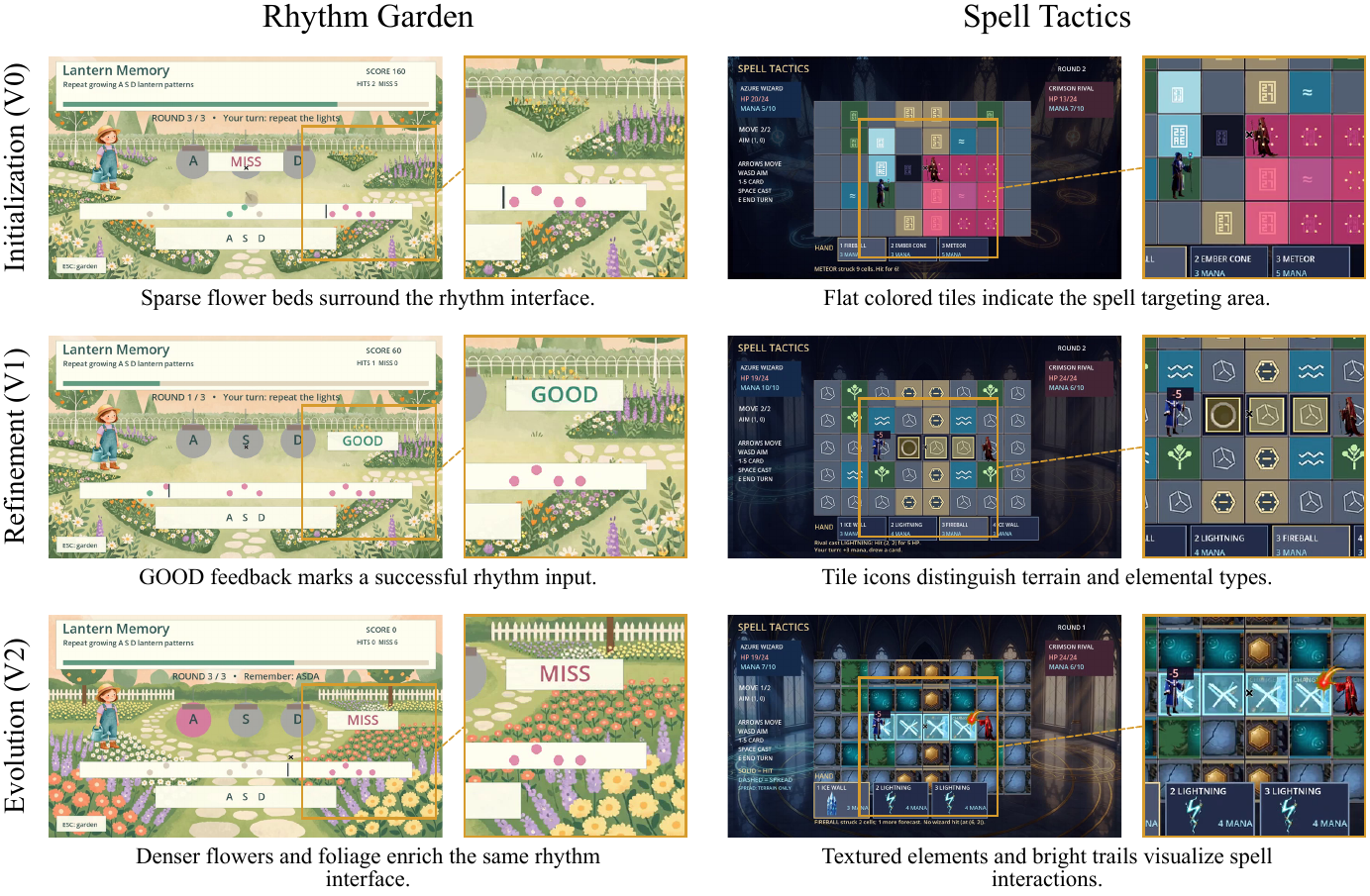}
  \caption{Rhythm Garden and Spell Tactics across \textrm{V0}--\textrm{V2}, with each full frame paired with an enlarged detail. The examples highlight richer scenery and more detailed terrain and spell effects.}
  \label{fig:supp-stage-godot-b}
\end{figure}

\begin{figure}[htbp]

  \centering
  \captionsetup{skip=2pt}
  \includegraphics[width=0.95\textwidth]{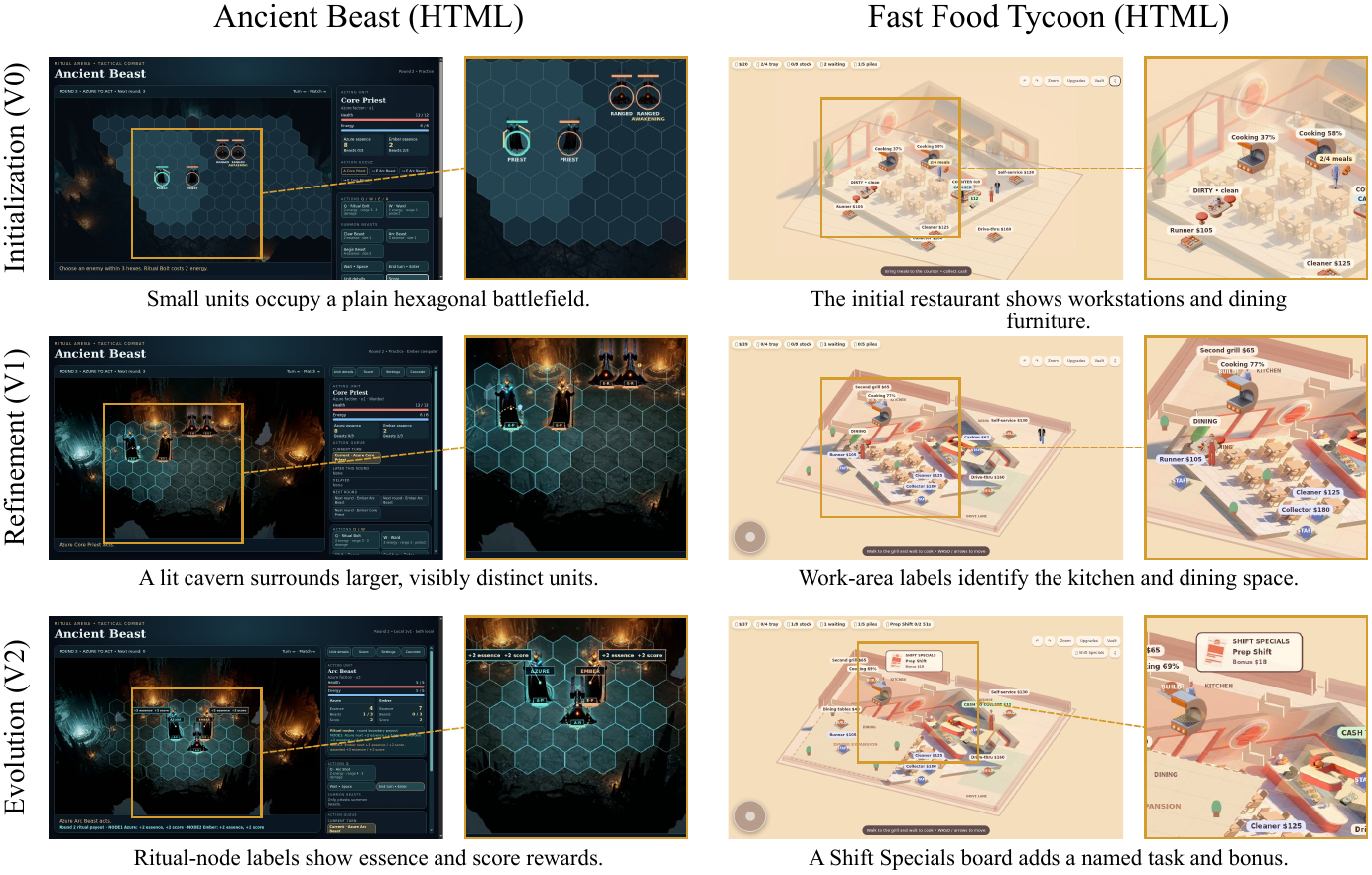}
  \caption{Ancient Beast and Fast Food Tycoon across \textrm{V0}--\textrm{V2}, with each full frame paired with an enlarged detail. The highlighted regions show revised game environments, ritual-node rewards, and a restaurant shift-contract interface.}
  \label{fig:supp-stage-html}
\end{figure}

Fig.~\ref{fig:supp-stage-godot-a}--\ref{fig:supp-stage-html} extend the main-paper comparison with 4 Godot games and 2 HTML games. Each game uses one recorded frame per checkpoint, paired with an enlarged view of the yellow-boxed region. The rows show Initialization (\textrm{V0}), Refinement (\textrm{V1}), and Evolution (\textrm{V2}). Short descriptions identify visible changes to buildings, terrain, visual feedback, and gameplay interfaces. Source frames and crop coordinates are recorded with the figure assets. Play routes and states can differ across checkpoints, so these examples illustrate visible changes rather than matched-input quality gains.

\section{Scoring Issues in Game-Generation Benchmarks}
\label{sec:scoring}
GameCraft-Bench and GameASG-Bench provide common evaluation procedures, but their scores cover different aspects of game quality. We examine how demonstration selection, visual sampling, and functional checks affect the evidence available to their evaluators. The analysis centers on a Boxing Gym case and 2 GameASG tasks, with additional Dungeon Guild and Bird examples. These cases identify specific limitations without estimating how often they occur across all tasks.

\subsection{GameCraft-Bench}

\paragraph{Demo selection affects scores.}
GameCraft-Bench~\cite{luo2026gamecraft} scores gameplay shown by submitted demonstrations. In the Boxing Gym case, we hold the per-demo judgments fixed and recompute all 127 nonempty subsets of 7 demos. Overall ranges from 20.6 to 60.7, compared with 59.1 for the full submission. Of these subsets, 11 score higher than the full submission. Repeating this aggregation analysis with judgments from another model gives a range of 16.9--59.4, with 12 subsets exceeding the full-submission score. These are recombinations of fixed judgments, not repeated evaluations of independently generated games.

In the examined task, functional items take the maximum over demos, while art items take the mean. The jump-rope demo receives 0.60, 0.50, and 0.30 for palette, characters, and animation. Fight demos receive about 0.80, 0.60, and 0.55 on these items. Removing the 2 training demos raises Overall from 59.1 to 60.7 without changing the game. Lower ratings for weaker scenes are reasonable, but selecting which scenes enter the average changes the resulting score. \textbf{This case shows that submission selection can increase a score without improving the underlying game.}

\paragraph{Local demonstrations do not establish connected progression.}
Scenario-specific launches can expose mechanics without showing how players reach them. In the boxing recordings, all 3 opponents are loaded directly. Both demos launched from the title screen stop in round 1, leaving advancement after a win unobserved. The opponent-variety item receives 0.40, with the rationale stating that only one opponent is shown. The local clips therefore do not establish that normal play connects the initial state to later opponents. This distinction motivates checking both scenario coverage and progression through ordinary gameplay.

\paragraph{Replay behavior depends on execution conditions.}
In the examined boxing implementation, simulation speed depends on frame rate. It runs at about 6.5$\times$ normal speed on an idle machine and 0.6$\times$ under evaluation load. An input sequence that wins by KO during development misses every punch during evaluation. This is a timing defect in the game as well as a source of variation in the displayed evidence. A low score for an unobserved KO cannot by itself distinguish a missing mechanic from an input sequence that no longer reaches it. The trajectory module in Sec.~\ref{sec:supp-rsi-showcase} addresses changed input requirements on frozen projects, without using benchmark scores to choose routes.

\paragraph{Trajectory preparation on a frozen game.}
\label{sec:supp-trace-sensitivity}
Tab.~\ref{tab:demo-sensitivity} gives a controlled Dungeon Guild pilot. The initialization game scores 54.59 with its original 9 traces. The refined game scores 49.42 when replayed with those traces, but 61.06 with 9 regenerated traces. The 2 refined evaluations use identical project files, rubric, and judge model. The full quest loop criterion V1 rises from 0 to 1 with regenerated traces. Thus, the old traces failed to expose a relevant path in the refined game. Each row uses 9 demos and 1 GPT-6 Sol judgment pass. Here, criterion V1 tests the connected quest loop and is distinct from the refined checkpoint label. These are single judge passes, so some of the score difference may also reflect judgment variability.

\begin{table}[!htbp]
  \centering
  \small
  \setlength{\tabcolsep}{5pt}
  \caption{Dungeon Guild pilot with original and regenerated demo traces. The same refined project receives scores differing by 11.64 points under the 2 trace sets.}
  \label{tab:demo-sensitivity}
  \begin{tabular}{lcc}
    \hline
    Game and traces & Overall & Quest-loop criterion \\
    \hline
    \textrm{V0}, original & 54.59 & 1 \\
    \textrm{V1}, original & 49.42 & 0 \\
    \textrm{V1}, regenerated & 61.06 & 1 \\
    \hline
  \end{tabular}
\end{table}

\paragraph{A mechanic-focused trace can lower mean art scores.}
In a frozen Bird pilot, a route-focused demonstration provides the highest flight-mechanic score, M1=1.0. Its A1 and A4 scores are both 0.5. The A4 rationale notes that the result panel and retry button are absent from this clip. Adding the clip to the other 5 raises M1 from 0.5 to 1.0, while the mean A score falls from 0.9375 to 0.9063. Overall still rises from 69.00 to 74.06 on the 0--100 scale. This recomputation reuses 1 set of per-demo judgments and does not rerun the judge. A trace that does not show the result screen cannot assess its UI styling. Explicit visibility annotations could help distinguish absent evidence from poor appearance, but their effect on scoring requires evaluation.

\paragraph{Art scores and player preference.}
The boxing art rubric includes a title screen with a working start button. The examined game receives 0.95 on this criterion despite using geometric character drawings. This item mixes functional completeness with visual presentation, so its score alone does not establish aesthetic appeal. Other judgments show imperfect agreement with explicit style rules. The evaluator describes 4 demos as not using a warm palette but assigns palette scores of 0.60--0.80. It also assigns nonzero scores where the rubric specifies zero for flat geometric shapes or hard cuts. These observations concern the examined rubric items and judgments.

The 2 judge models have a per-item correlation of 0.945 in this boxing case. This measures agreement between their scores on these items, rather than repeatability across independent judging runs or agreement with player preference. Human agreement when applying the same rubric likewise addresses rubric interpretation. Establishing whether higher scores predict preferred games requires separate preference evidence. Our clip-based human study in Sec.~\ref{sec:supp-human-stages} provides a separate assessment of perceived quality and willingness to play.

\begin{figure}[!htbp]
  \centering
  \includegraphics[width=0.9\linewidth,height=0.72\textheight,keepaspectratio]{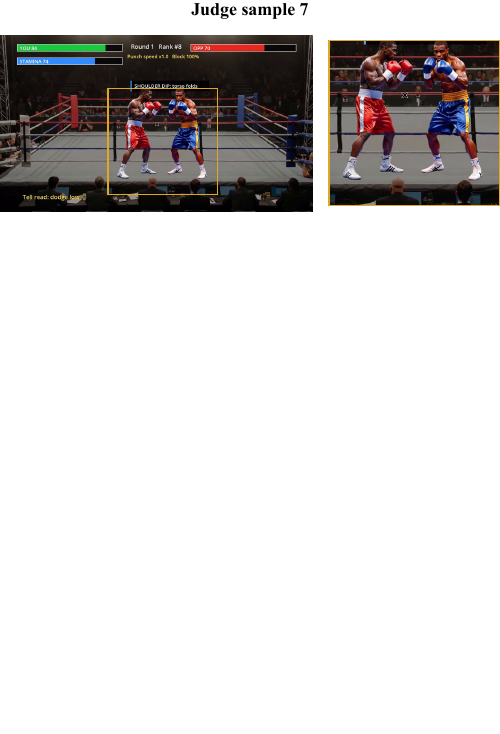}
  \caption{A brief crouching action between 2 judge samples in our Boxing Gym replay, with full frames on the left and enlarged details on the right. The middle row reveals an intermediate pose absent from the adjacent samples, illustrating incomplete temporal evidence from sparse sampling.}
  \label{fig:transient}
\end{figure}

\paragraph{Sparse sampling can miss brief visual events.}
Fig.~\ref{fig:transient} illustrates a brief action in our Boxing Gym output using the recorded video and the images supplied to the judge. The player is upright in judge samples 7 and 8, but crouches briefly between them at 3.467\,s in the video. The adjacent samples therefore omit this intermediate pose. This is a normal gameplay action, not a visual defect. The example shows that sampled still frames can leave the action sequence incomplete, even when later frames retain its state changes.

The earlier boxing audit also reveals problems at other points in evaluation. Hard cuts appear in sampled evidence and are mentioned by the judge, but do not receive the zero scores specified by the rubric. A nearly transparent KO banner and a jab-only strategy that wins in about 10\,s are absent from the submitted demonstrations. \textbf{Together, these examples distinguish missing gameplay coverage, missed temporal events, and inconsistent rubric application.} Improving one of these does not automatically resolve the others.

\subsection{GameASG-Bench}

\begin{figure}[!htbp]
  \centering
  \includegraphics[width=\linewidth]{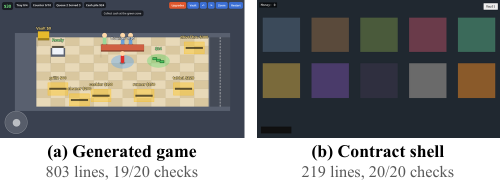}
  \caption{A generated \texttt{fast-food-tycoon} and a diagnostic contract shell. The shell passes the task's checks despite lacking the depicted gameplay content.}
  \label{fig:shell}
\end{figure}

\begin{figure}[!t]
  \centering
  \includegraphics[width=\textwidth]{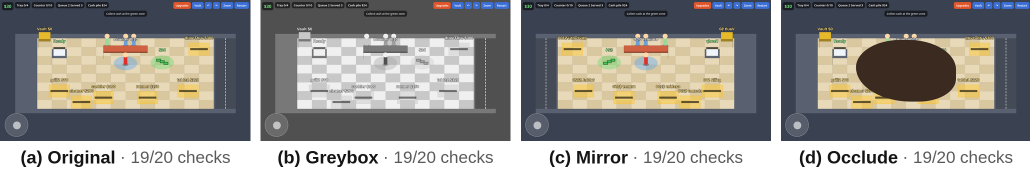}
  \caption{Visual variants of \texttt{fast-food-tycoon} in the same game state. These variants retain their original scores, and the original and 3 variants of \texttt{asmr-sparkle-shine} all pass 19/19 checks.}
  \label{fig:visual}
\end{figure}

\paragraph{Mean pass rates and complete task success differ.}
GameASG-Bench~\cite{zhang2026gameasg} reports a best mean L2 of 93.2\%, while 55.3\% of tasks (26 of 47) pass all required checks. These metrics answer different questions: the fraction of checks passed and whether every required condition holds. In an initial evaluation of our 2 example games, browser startup timeouts caused checks to fail despite a mean of 87.2. Both games passed the required checks after the environment was corrected. This example also distinguishes an execution-environment problem from an implemented-game defect.

Checks marked not applicable are excluded from their denominators. In 3 examined tasks, applicability of required checks depends on state reported by the game. For example, the \texttt{tap-tap-shots} economy check is marked not applicable when the game reports that its shop scenario cannot load. Aggregate scores should therefore be interpreted alongside strict success and the set of applicable checks.

\paragraph{Reported states need not establish playable content.}
Using the public specification and check code, we constructed a 219-line \texttt{fast-food-tycoon} contract shell as an evaluation diagnostic. Its display contains 10 fixed colored blocks and a moving bar, without a player, food, counter, or customers. The expected game content exists only in states returned to the tests. This shell passes all 20 checks, whereas the normally generated 803-line game passes 19 (Fig.~\ref{fig:shell}). The shell is a diagnostic counterexample, not an output used in the main method comparison.

Most checks in this task inspect reported states, including coordinates used to assess movement. The image checks establish that a screenshot is nonblank and changes over time. The shell's moving bar satisfies the latter without implementing the intended gameplay. Although the specification requires reported states to match visible interactions, these checks do not establish that correspondence in this example. \textbf{Passing the examined test suite is insufficient to verify that players can see and interact with the reported content.}

\paragraph{Visual changes can leave functional scores unchanged.}
GameASG-Bench has no separate art score. For each of 2 evaluated games, we construct 3 variants that change presentation while preserving game logic and reported states. A greybox variant maps colors to 6 gray levels and removes shadows. A mirror variant flips the display horizontally, reversing the apparent direction of movement. An occlusion variant covers the center of the play area. All 6 variants retain their respective original scores and the same per-check outcomes (Fig.~\ref{fig:visual}).

The tested changes preserve nonblank, changing screenshots, so they do not violate the image checks despite substantial visual differences. Mirroring and occlusion also affect how players perceive and control the game. These controlled examples demonstrate insensitivity to the tested transformations. They support the interpretation in Sec.~\ref{sec:supp-gameasg-limits} that visual gains can coexist with unchanged functional scores. They do not establish that every visual change is undetectable or that this mechanism explains every unchanged score between Refinement and Evolution.

\paragraph{Fixed checks can miss player-facing defects.}
Playtesting the 2 example games identifies 19 problems not detected by their existing checks. In \texttt{fast-food-tycoon}, per-frame rounding prevents the vault balance from growing, while the check only requires a nonnegative balance. Another check collects food through a test-specific movement command, whereas ordinary play cannot pick up food spawned outside the grill's pickup range. In \texttt{asmr-sparkle-shine}, foam obscures the cleaned object and tool icons render as boxes, but the check only verifies a changing cleanliness value. These examples concern loose predicates, differences between test and player routes, and missing visual checks.

GameASG's independently executed tests provide a common functional comparison without using our demonstration selection procedure. Their shared execution does not make them exhaustive measures of player experience. \textbf{The examples motivate combining functional tests with observed gameplay and human ratings.} They do not invalidate the functional properties that the tests directly verify.

\section{Main Comparison Protocol}
\label{sec:supp-main-protocol}
We evaluate all 140 GameCraft-Bench tasks across 15 categories and all 47 GameASG-Bench tasks across 12 categories. Within each benchmark, all compared methods and checkpoints use the same task set. We denote the games after Initialization, Refinement, and the first Evolution cycle by \textrm{V0}, \textrm{V1}, and \textrm{V2}, respectively.

Adapted Play2Code and Play2Evolve start from identical Initialization packages and use the same agent backbone and video generator. Play2Code directly diagnoses gameplay issues and can revise code or call MiniMax H3 to generate materials. For a fair comparison, both methods receive the same total development-token budget. Within each benchmark, evaluation uses the same benchmark version and evaluator configuration for all compared methods and checkpoints. Benchmark judgments are withheld from development.

\paragraph{GameASG-Bench evaluation.}
GameASG-Bench evaluates required game functionality through its own testing procedure. The benchmark executes the prescribed checks independently of our development Player, rather than relying on gameplay trajectories prepared by our method. We submit frozen projects to the same evaluation procedure across methods and checkpoints. L1 and L2 Mean average per-game pass rates, while L2 P0/P1/P2 pool applicable checks across games. Strict task success requires valid delivery, completed evaluation, all L1 checks, and all applicable L2 P0/P1 checks to pass.

\paragraph{GameCraft-Bench evaluation.}
GameCraft-Bench evaluates gameplay exposed by submitted demonstration traces. Its scores therefore depend on both the implemented game and the interactions shown during evaluation. We freeze each project before preparing demonstrations and use the same demonstration budget and scoring configuration across compared methods. The trajectory comparison protocol in Sec.~\ref{sec:supp-rsi-showcase} preserves shared evaluation targets while allowing inputs to accommodate changes in game timing and navigation. Benchmark scores are not used to select or revise trajectories. We average per-game scores over the common task set. Overall averages the official task-weighted game scores rather than taking an unweighted mean of the 4 category columns.

\paragraph{What GameASG-Bench captures.}
\label{sec:supp-gameasg-limits}
GameASG-Bench emphasizes required implementation and prescribed runtime interactions, without a separate score for visual aesthetics. Evolution can add optional mechanics, richer motion, and scene content without changing these checks. This helps explain why aggregate strict success and mean L2 remain unchanged from \textrm{V1} to \textrm{V2}, despite improvements in visual appeal and willingness to play in the human evaluation.

\subsection{Development token accounting}
\label{sec:supp-token-accounting}
The development-token values for our checkpoints report the additional cost of each stage, averaged per game in millions. Initialization, Refinement, and the first Evolution cycle use 0.30M, 3.69M, and 4.89M tokens, respectively. Each stage's count excludes consumption from earlier stages. Play2Code uses 8.89M tokens per game under the matched total development-token budget.

Development accounting covers construction, planning, playtesting, internal judging, and code editing. Independent benchmark evaluation and post-development demonstration preparation are excluded. Video-generation compute and GPU time are separate from language-model token consumption. We apply the same development-token accounting convention to Play2Code and Play2Evolve.

\section{Behavior Skill Evaluation and Examples}
\label{sec:supp-behavior-skill}
We compare play programs with and without Behavior Skill on the same 10 frozen games, using identical scenario goals, Player settings, and action budgets. Skills are saved from earlier playtests and fixed before evaluation. Coverage, breadth, and depth follow Sec.~\ref{sec:supp-playtest-metrics}. This comparison measures exploration rather than downstream game quality, and reported playtest costs exclude prior Skill creation.

Historical examples illustrate the guidance: Dungeon Shop instructions direct the Player from the shop into the dungeon, while Wave Commander instructions encourage advancing beyond the shop to later waves. The latter reaches wave 4 instead of wave 2, but both reach the same predefined depth milestone. Skill does not always help, as an Archery Quest pair completes fewer stages after an update. These cases differ in inherited context and illustrate possible effects rather than isolating Skill's causal contribution.

\section{Human Evaluation Protocol with Play2Code}
\label{sec:supp-human-stages}
The main human study uses 30 selected games and 12 participants to compare adapted Play2Code with our Initialization, Refinement, and Evolution games. Participants view anonymous clips and score visual appeal, motion quality, absence of disruptive artifacts, willingness to play, and overall quality on 5-point scales. Willingness to play asks whether the viewer would want to try the game after watching the clip. Overall quality is a separate impression of the clip. We record clip duration, resolution, frame rate, and shown interaction for each condition. An optional overall ranking and free-text comment provide secondary evidence.

\begin{figure}[htbp]
  \centering
  \includegraphics[width=0.80\textwidth]{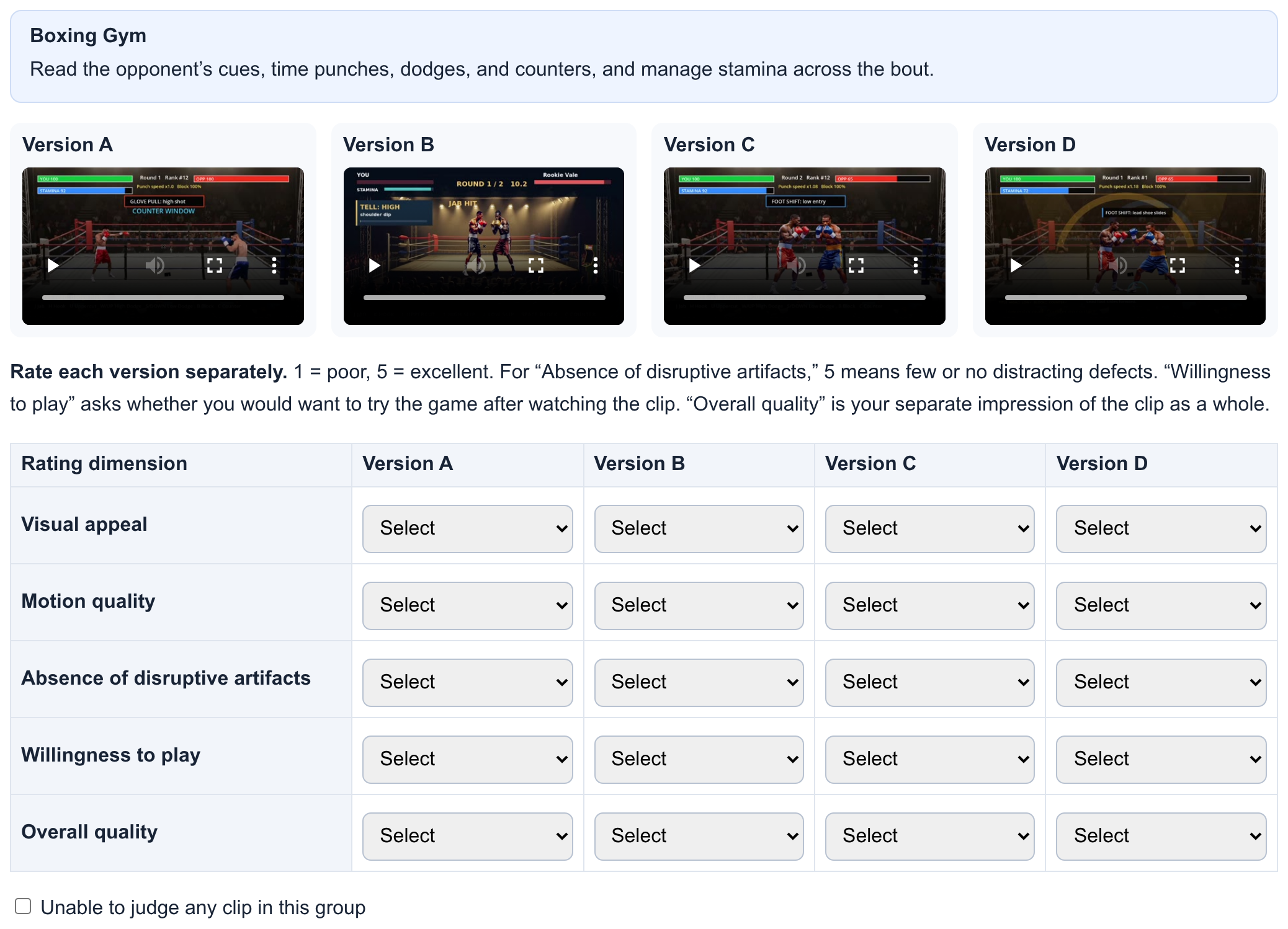}
  \caption{An example questionnaire screen from the archived English pilot, showing Boxing Gym clips from Play2Code and our Initialization, Refinement, and Evolution checkpoints. Anonymous version labels accompany the game rules and 5 separate rating dimensions.}
  \label{fig:supp-human-questionnaire}
  
\end{figure}

Fig.~\ref{fig:supp-human-questionnaire} illustrates the questionnaire interface with a 4-condition example from the archived English pilot. The screenshot is included only to document the interface. It is not used to compute the final human-evaluation means.

The rating anchors are as follows. Visual appeal ranges from 1 (unappealing) to 5 (highly appealing). Motion quality ranges from 1 (broken or incoherent motion) to 5 (clear, coherent motion). Artifact absence ranges from 1 (severe defects that disrupt viewing) to 5 (no noticeable disruptive defects). Willingness to play ranges from 1 (would not want to try) to 5 (strongly want to try). Overall quality ranges from 1 (very poor) to 5 (excellent) based on the clip as a whole. Each item also offers ``unable to judge.'' We do not replace these missing ratings with neutral scores. Participants may give the same score to multiple versions.

For each dimension and condition, we average valid ratings within a game, then average the matched game means so that every game has equal weight. ``Unable to judge'' responses are excluded from that dimension. We preserve the pairing of games and participants when computing condition differences. We summarize optional rankings separately by first-place share and pairwise preference.

\section{Failure Cases and Limitations}
\label{sec:discussion}
\label{sec:failure-cases}
\label{sec:supp-failure-analysis}

\paragraph{Incomplete exploration.}
A Player can trigger target events while missing later gameplay. In a Plant Defense pilot, similar event coverage concealed different progression through the game. Playtest feedback therefore depends on the routes explored, and unobserved content cannot be reliably diagnosed.

\paragraph{Unreliable execution or visual evidence.}
Recording and Player errors can obscure the game being evaluated. A Bird replay showed placeholder graphics until textures were imported, while an invalid key sequence interrupted a Hotline Heist playtest. These are evidence or execution failures, which must be distinguished from defects in game mechanics.

\paragraph{Evolution does not always yield a better candidate.}
In a Bird pilot, the evolved candidate reached less progress in the observed rollouts, and the final review retained the previous game. Different play policies limit attribution of this difference to the revision itself. Candidate rejection can avoid adopting an unconvincing change, but observed acceptance alone does not guarantee improvement across all gameplay paths.

\paragraph{Scope of evaluation and deployment.}
\label{sec:supp-craft-limits}
The benchmark cases in Sec.~\ref{sec:scoring} identify specific scoring limitations rather than their prevalence across all games. Our human study measures perceived quality and willingness to play from clips, rather than enjoyment after direct play or sustained engagement. Video regeneration also incurs additional compute, and transferring the system beyond the tested engines and generator requires further evaluation.

\end{document}